\documentclass[11pt]{article}

\usepackage[utf8]{inputenc}
\usepackage[T1]{fontenc}
\usepackage{hyperref}
\usepackage{url}
\usepackage{booktabs}
\usepackage{amsfonts}
\usepackage{amsmath}
\usepackage{nicefrac}
\usepackage[expansion=false]{microtype}
\usepackage{xcolor}
\usepackage{graphicx}
\usepackage{subcaption}
\usepackage{multirow}
\usepackage[margin=1in]{geometry}
\usepackage{listings}

\usepackage{cite}

\lstdefinestyle{promptstyle}{
  basicstyle=\footnotesize\ttfamily,
  breaklines=true,
  breakatwhitespace=true,
  breakindent=0pt,
  columns=flexible,
  keepspaces=true,
  showstringspaces=false,
  frame=none,
  xleftmargin=1em,
  xrightmargin=1em,
  aboveskip=0.5em,
  belowskip=0.5em
}

\lstnewenvironment{promptbox}
  {\lstset{style=promptstyle}}
  {}

\title{\textbf{CRAFT: Continuous Reasoning and Agentic Feedback Tuning \\
for Multimodal Text-to-Image Generation}}

\author{
  V. Kovalev\thanks{Corresponding authors: \texttt{\{valentin, alexander, denis, alexb, dima\}@flymy.ai}}$^{1}$ \and
  A. Kuvshinov$^{1}$ \and
  A. Buzovkin$^{1}$ \and
  D. Pokidov$^{1}$ \and
  D. Timonin$^{1}$
}

\date{}

\begin{document}

\maketitle

% Abstract
%==============================================================================
% Abstract
%==============================================================================

\begin{abstract}
Recent work has shown that inference-time reasoning and reflection can improve
text-to-image generation without retraining. However, existing approaches often
rely on implicit, holistic critiques or unconstrained prompt rewrites, making
their behavior difficult to interpret, control, or stop reliably. In contrast,
large language models have benefited from explicit, structured forms of
\emph{thinking} based on verification, targeted correction, and early stopping.

We introduce \textbf{CRAFT} (\textbf{C}ontinuous \textbf{R}easoning and
\textbf{A}gentic \textbf{F}eedback \textbf{T}uning), a training-free,
model-agnostic framework that brings this structured reasoning paradigm to
multimodal image generation. CRAFT decomposes a prompt into
dependency-structured visual questions, verifies generated images using a
vision-language model, and applies targeted prompt edits through an LLM agent
only where constraints fail. The process iterates with an explicit stopping
criterion once all constraints are satisfied, yielding an interpretable and
controllable inference-time refinement loop.

Across multiple model families and challenging benchmarks, CRAFT consistently
improves compositional accuracy, text rendering, and preference-based
evaluations, with particularly strong gains for lightweight generators.
Importantly, these improvements incur only a negligible inference-time overhead,
allowing smaller or cheaper models to approach the quality of substantially
more expensive systems. Our results suggest that explicitly structured,
constraint-driven inference-time reasoning is a key ingredient for improving
the reliability of multimodal generative models.
\end{abstract}

% Main sections
%==============================================================================
% Section: Introduction
%==============================================================================

\section{Introduction}
\label{sec:introduction}

Recent progress in large language models has demonstrated that explicit
inference-time reasoning--including verification, reflection, and targeted
correction--can substantially improve reliability without additional training.
Rather than relying solely on larger models or additional data, these systems
benefit from structured forms of \emph{thinking} that make intermediate decisions
explicit, checkable, and correctable. Related ideas have recently begun to
influence multimodal generation, where reflection-based and agentic pipelines
show that post-hoc critique can improve text-to-image (T2I) outputs without
retraining~\cite{zhuo2025reflection,liu2025maestro}.

At the same time, modern T2I systems remain predominantly single-shot and
reactive. Despite impressive visual fidelity achieved by diffusion- and
transformer-based models~\cite{rombach2022high,ramesh2022hierarchical,
saharia2022photorealistic,podell2023sdxl,betker2023improving,yu2022scaling},
they often struggle with long, compositional, and text-heavy prompts, where
multiple constraints must be satisfied simultaneously. Generation is typically
controlled through a single textual channel--the prompt--which acts as the
primary mechanism for steering content, composition, style, and even safety in
black-box deployments.

While inference-time refinement can mitigate some of these issues, existing
approaches often rely on implicit, holistic critiques or unconstrained prompt
rewrites~\cite{pryzant2023automatic,zhuo2025reflection,liu2025maestro}, making
their behavior difficult to interpret, control, or stop reliably. A key
challenge is that image quality in open-domain T2I is inherently multi-criteria:
a satisfactory image must satisfy compositional constraints, maintain object
attributes and relations, adhere to style and layout, render text correctly, and
avoid visual artifacts--all at once.

Recent work therefore emphasizes explicit, question-based evaluation, where
prompts are decomposed into verifiable visual statements that better correlate
with human judgments~\cite{hu2023tifa,lin2024evaluating}. Benchmarks such as
Davidsonian Scene Graph (DSG) and GenEval further demonstrate that structured,
checkable constraints provide a more reliable foundation for evaluating and
improving T2I systems~\cite{cho2024davidsonian,ghosh2023geneval}. These insights
suggest that open-domain T2I models could similarly benefit from converting
implicit textual requirements into explicit, verifiable questions.

%==============================================================================
% Section: Related Work (Method Overview)
%==============================================================================

\section{Related Work}
\label{sec:related}

\begin{figure}[htbp]
\centering
\includegraphics[width=0.9\textwidth]{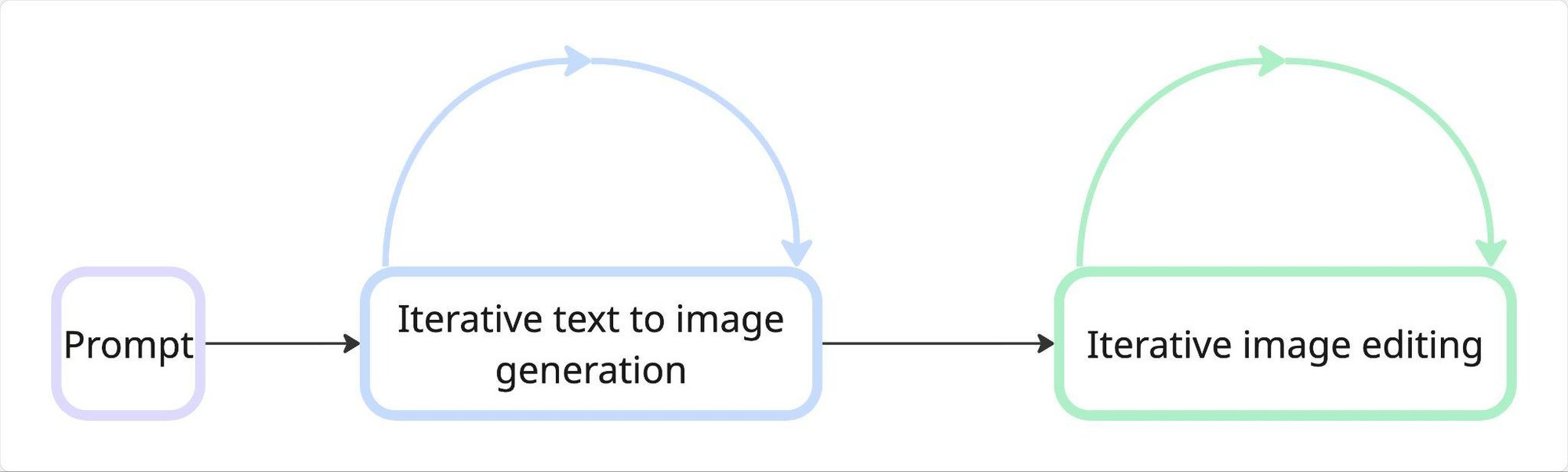}
\caption{Schema of CRAFT image generation and image editing.}
\label{fig:schema}
\end{figure}

We propose \textbf{CRAFT}, a training-free reasoning layer for text-to-image 
generation and image editing (Figure~\ref{fig:schema}) that performs iterative 
refinement while avoiding redundant steps once the image satisfies the target 
quality criteria. In our implementation, the text-to-image and image-editing 
components of CRAFT can be used either sequentially or independently, which 
allows the framework to be integrated flexibly into diverse generation pipelines 
and model stacks.

% Method overview figures

%==============================================================================
% Section: CRAFT T2I Generator
%==============================================================================

\section{CRAFT T2I Generator}
\label{sec:craft_t2i}

The generator works iteratively. Its goal is to turn the initial prompt into a 
controlled, unambiguous improved prompt and a best image under the DVQ criterion.

\begin{figure}[h!]
    \centering
    \includegraphics[width=0.85\textwidth]{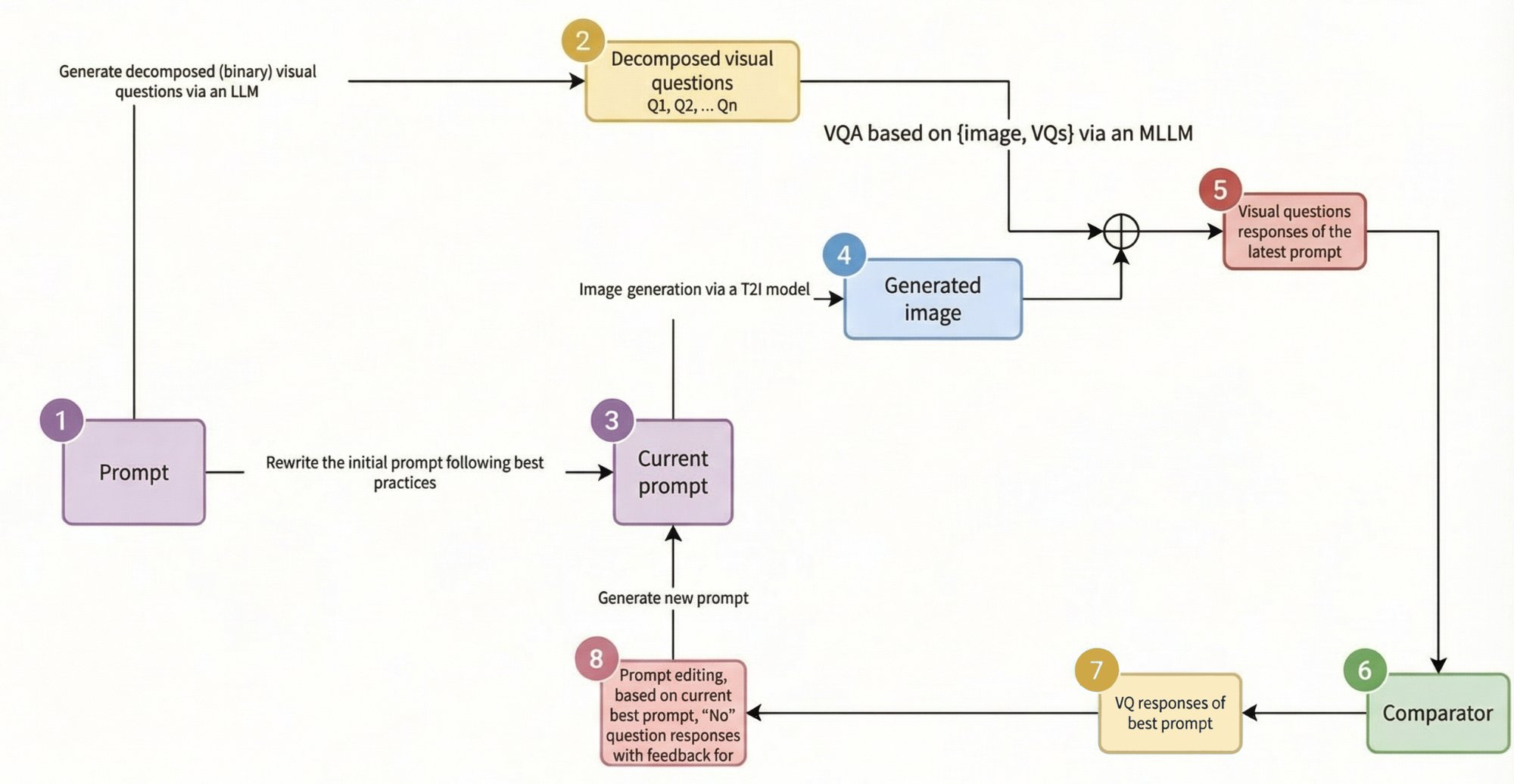}
    \caption{Schema of CRAFT T2I generator.}
    \label{fig:t2i_schema}
    \end{figure}

\paragraph{Workflow.}

%------------------------------------------------------------------------------
% Initialization
%------------------------------------------------------------------------------
\textbf{Initialization.} From the user prompt $\mathcal{P}_0$, two processes 
start in parallel:

%------------------------------------------------------------------------------
% Visual-question generation
%------------------------------------------------------------------------------
\textbf{Visual-question generation:} We construct a set of explicit visual 
questions $\mathcal{Q}$ that describe the intended scene and, when applicable, 
a dependency graph $\mathcal{G}$ so that prerequisite questions precede their 
dependents (e.g., ``Is there a cat?'' before ``Is the cat 
red?'')~\cite{cho2024davidsonian}. The base question set covers objects, 
attributes, spatial relations, and high-level scene properties derived from 
the user prompt. In addition, we augment $\mathcal{Q}$ with two generic 
diagnostic queries: a text-visibility question ``Is the text in the attached 
image fully clear and easily readable by a human?'' and an artifact check 
``Does the image look natural and free from visual glitches, artifacts, or 
distortions?'' Both are answered strictly with YES/NO to keep the signal 
discrete. The text-visibility question is included only when the prompt 
contains explicit textual content that is intended to appear in the image 
(e.g., spans in quotation marks or other text-to-render instructions). In such 
cases, the question acts as a global check on whether the rendered typography 
is legible, complementing the fine-grained DVQ questions about whether specific 
strings appear. The artifact check is always present and serves as a generic 
quality probe, capturing issues such as distorted text, warped geometry, or 
rendering glitches. Together, these additions extend the question set beyond 
purely semantic correctness to cover text legibility and low-level visual 
fidelity, which are then used as constraints in subsequent refinement steps.

%------------------------------------------------------------------------------
% Prompt rewrite
%------------------------------------------------------------------------------
\textbf{Prompt rewrite:} use an LLM to rewrite $\mathcal{P}_0$ following best 
practices for T2I prompting, reducing ambiguity while preserving the original 
intent~\cite{pryzant2023automatic}. At each iteration $t$, the system applies 
a controlled rewriting step to produce an improved prompt while preserving the 
user's intent. The rewriting module is implemented as a constrained LLM-based 
editor that operates under a fixed set of rules. First, the module enforces a 
canonical structure for text-to-image prompting -- explicitly ensuring that the 
rewritten prompt contains a well-formed subject description, 
contextual/environmental details, and a stylistic specification. If the original 
prompt does not specify a style, the model inserts a default realistic style. 
The editor also enhances semantic clarity by adding descriptive modifiers and 
resolving ambiguous phrasing, while avoiding stylistic drift beyond the bounds 
of the user's initial intent.

A key component of the rewriting logic is the explicit handling of textual 
elements. If the original prompt contains text enclosed in quotation marks 
(e.g., headers, labels, signage), this text is treated as literal content that 
must appear verbatim in the generated image. The rewriting module preserves 
such spans exactly -- including case, punctuation, and quotation marks -- and 
blocks their rephrasing, translation, or removal. In contrast, unquoted textual 
references (e.g., ``with the title $X$'', ``named $Y$'') are interpreted as 
metadata or thematic context rather than visible text; these are allowed to be 
paraphrased or reframed as stylistic descriptors. This distinction prevents 
both unintended text hallucination and accidental loss or corruption of required 
text elements.

%------------------------------------------------------------------------------
% Generation
%------------------------------------------------------------------------------
\textbf{Generation.} Using the updated prompt $\mathcal{P}_t$, the system 
generates an image $\mathcal{I}_t$ with a text-to-image (T2I) model 
$\mathcal{M}_{\text{T2I}}$:
\begin{equation}
\mathcal{I}_t = \mathcal{M}_{\text{T2I}}(\mathcal{P}_t; \theta_t)
\end{equation}
where $\theta_t$ represents deterministic sampling parameters tied to iteration 
index. The framework is model-agnostic: any black-box T2I generator can be 
plugged into this stage without architectural changes, including diffusion-based 
models (e.g., FLUX, SD derivatives), autoregressive generators, or proprietary 
APIs.

%------------------------------------------------------------------------------
% DVQ evaluation
%------------------------------------------------------------------------------
\textbf{DVQ evaluation with rationales.} Given $\mathcal{I}_t$ and 
$\mathcal{Q}$, an MLLM $\mathcal{M}_{\text{VLM}}$ answers each question 
$q \in \mathcal{Q}$ with $\{$YES, NO$\}$. For every NO, it also produces a 
short rationale $r_q$ explaining the failure:
\begin{equation}
\text{score}_{\text{DVQ}}(\mathcal{I}_t) = \frac{1}{|\mathcal{Q}|} 
\sum_{q \in \mathcal{Q}} \mathbb{1}[\mathcal{M}_{\text{VLM}}(\mathcal{I}_t, q) 
= \text{YES}]
\end{equation}
These rationales guide the next prompt edits~\cite{zhuo2025reflection} for 
better generation results.

%------------------------------------------------------------------------------
% Selection
%------------------------------------------------------------------------------
\textbf{Selection (iterative update).}
\begin{itemize}
\item \textbf{First iteration:} skip comparison and set the incumbent best to 
the current image and prompt; store the DVQ answers and rationales.
\item \textbf{Later iterations:} compare the newly generated image against the 
incumbent using the DVQ score $\text{score}_{\text{DVQ}}(\mathcal{I}_t)$. If 
the new image yields a higher score, it becomes the new incumbent:
\begin{equation}
\mathcal{I}_{\text{best}} = \begin{cases}
\mathcal{I}_t & \text{if } \text{score}_{\text{DVQ}}(\mathcal{I}_t) > 
\text{score}_{\text{DVQ}}(\mathcal{I}_{\text{best}}) \\
\text{Judge}(\mathcal{I}_t, \mathcal{I}_{\text{best}}, \mathcal{P}_0) & 
\text{if scores are equal}
\end{cases}
\end{equation}
When the YES counts are equal, we invoke an MLLM-based side-by-side comparison. 
The judge selects the image that better satisfies the prompt and exhibits fewer 
visual artifacts.
\end{itemize}

%------------------------------------------------------------------------------
% Targeted prompt update
%------------------------------------------------------------------------------
\textbf{Targeted prompt update.} Provide the initial prompt $\mathcal{P}_0$, 
the current prompt $\mathcal{P}_t$, the question set $\mathcal{Q}$, and all DVQ 
answers and rationales to an LLM $\mathcal{M}_{\text{LLM}}$. The LLM produces 
a targeted revision:
\begin{equation}
\mathcal{P}_{t+1} = \mathcal{M}_{\text{LLM}}(\mathcal{P}_0, \mathcal{P}_t, 
\mathcal{Q}, \{r_q : q \in \mathcal{Q}_{\text{failed}}\})
\end{equation}
that aims to increase the YES rate on $\mathcal{Q}$ while preserving attributes 
from $\mathcal{P}_0$.

%------------------------------------------------------------------------------
% Stopping criterion
%------------------------------------------------------------------------------
\textbf{Stopping.} Run for a configurable maximum number of iterations 
$T_{\max}$, or stop early when:
\begin{equation}
\text{score}_{\text{DVQ}}(\mathcal{I}_t) = 1.0 \quad \text{(all questions 
answered YES)}
\end{equation}

%==============================================================================
% Section: CRAFT Image Editing
%==============================================================================

\section{CRAFT Image Editing}
\label{sec:craft_editing}

\begin{figure}[h!]
\centering
\includegraphics[width=0.7\textwidth]{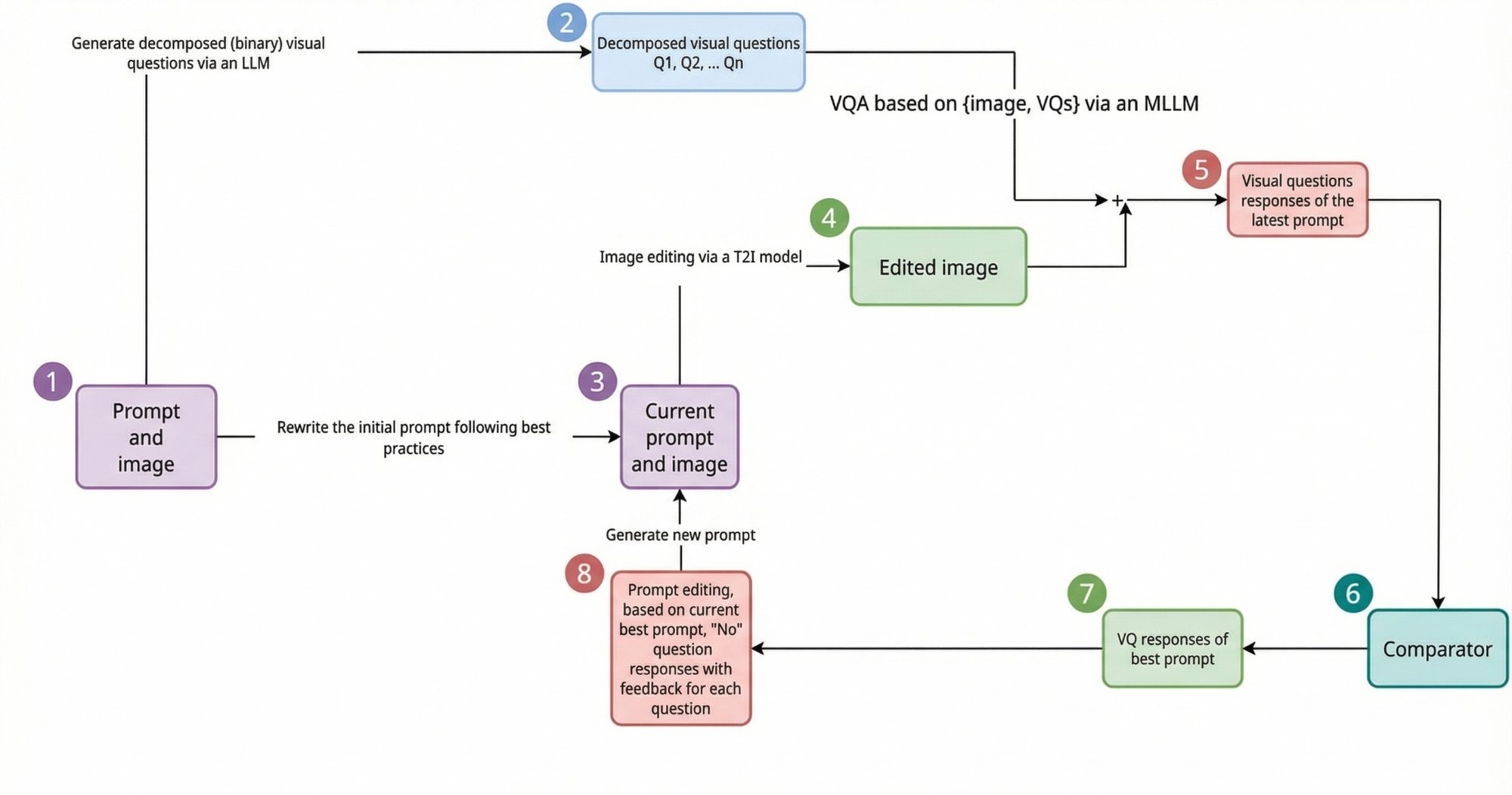}
\caption{CRAFT image editing schema.}
\label{fig:editing_schema}
\end{figure}

CRAFT iterative image editing (Figure~\ref{fig:editing_schema}) follows the 
same procedure as the iterative image generation loop, with two adaptations: 
(i) the prompt-rewriting and task prompts are tailored for editing, and (ii) 
the inputs consist of both a prompt and an image. The module can run 
independently or be composed with the iterative image generator.

%==============================================================================
% Section: Dataset and Models
%==============================================================================

\section{Dataset and Models}
\label{sec:dataset_and_models}

We evaluated using DSG and DVQ metrics~\cite{cho2024davidsonian}. In addition 
to the main 2-iteration configuration, we tested multiple backbone models and 
multiple iteration budgets for both generation and editing (see Model 
configurations for baseline and CRAFT). For each configuration, we computed all 
metrics across several datasets to assess robustness and cross-model consistency. 
For text-to-image (T2I), we ran a 2-iteration setup with the Qwen-Image 
model~\cite{wu2025qwen}. For image editing, we used a 2-iteration setup with 
Gemini 2.5 Flash by Google. We further included two preference-based metrics 
inspired by the Maestro paper -- auto side-by-side and auto side-by-side 
advantage~\cite{liu2025maestro}. As evaluation data, we used 
DSG-1K~\cite{cho2024davidsonian} and the Parti-Prompt (P2) 
suite~\cite{yu2022scaling}. We also report results on the P2-hard subset of
Parti-Prompt, following the Maestro protocol. Results are presented below
(Table~\ref{tab:dsg1k}, Table~\ref{tab:parti}).

%------------------------------------------------------------------------------
% Model configurations
%------------------------------------------------------------------------------
\subsection{Model Configurations for Baseline and CRAFT}
\label{sec:models}

For our evaluations, we consider five families of T2I models. For each model, 
we report results in two configurations: a baseline setting and a corresponding 
CRAFT setting.

\paragraph{Baseline 1 (B1) -- FLUX-Schnell:} Images are generated with FLUX-Schnell 
using a single forward pass and standard generation parameters.

\paragraph{CRAFT 1 (C1) -- FLUX-Schnell (2 iterations, no image editing):} We apply 
the full CRAFT pipeline for two iterations, using FLUX-Schnell for all 
generations, without any image-editing model. Only prompt-level improvements 
and DVQ-guided refinement are used.

\paragraph{Baseline 2 (B2) -- FLUX-Dev:} Images are generated with FLUX-Dev using a 
single forward pass and standard generation parameters.

\paragraph{CRAFT 2 (C2) -- FLUX-Dev (2 iterations, no image editing):} We apply the 
full CRAFT pipeline for two iterations, using FLUX-Dev for all generations, 
without any image-editing model.

\paragraph{Baseline 3 (B3) -- Qwen-Image (1 iteration):} Images are generated with 
Qwen-Image using default settings, without iterative refinement.

\paragraph{CRAFT 3 (C3) -- Qwen-Image (2 T2I iterations) + Gemini 2.5 Flash (2 
editing iterations):} We run two CRAFT iterations with Qwen-Image, followed by 
two DVQ-guided image-editing iterations using Gemini 2.5 Flash.

\paragraph{Baseline 4 (B4) -- Z-Image-Turbo (1 iteration):} Images are generated 
with Z-Image-Turbo using default settings and a single forward pass.

\paragraph{CRAFT 4 (C4) -- Z-Image-Turbo (2 T2I iterations):} We run two DVQ-guided 
CRAFT iterations with Z-Image-Turbo.

\paragraph{Baseline 5 (B5) -- FLUX-2 Pro (1 iteration):} Images are generated with 
FLUX-2 Pro using default parameters.

\paragraph{CRAFT 5 (C5) -- FLUX-2 Pro (2 T2I iterations):} We run two CRAFT 
iterations with FLUX-2 Pro, using DVQ feedback to iteratively refine the prompt.

%==============================================================================
% Section: Implementation Details
%==============================================================================

\section{Implementation Details}
\label{sec:implementation}

This section describes the specific models, hyperparameters, and system 
configurations used in our CRAFT implementation.

%------------------------------------------------------------------------------
% Model Selection
%------------------------------------------------------------------------------
\subsection{Model Selection}
\label{sec:model_selection}

\paragraph{Vision-Language Model (VLM):}
All VQA evaluation, DVQ question answering, image critique, and side-by-side 
comparison steps use \textbf{GPT-5-nano} (gpt-5-nano-2025-08-07) via OpenAI 
API. We selected this model for its balance of cost efficiency, fast inference 
speed, and strong vision-understanding capabilities. The model demonstrates 
robust performance on binary visual question answering tasks and provides 
consistent, structured outputs suitable for automated evaluation pipelines.

\paragraph{Language Model (LLM):}
Prompt rewriting, DVQ question generation, and targeted refinement operations 
use \textbf{GPT-5-nano} (gpt-5-nano-2025-08-07). The model's strong 
instruction-following capabilities and ability to maintain semantic consistency 
across iterative edits make it well-suited for constrained prompt engineering 
tasks.

\paragraph{Text-to-Image Backbones:}
We evaluate CRAFT with five T2I models representing different performance and 
cost tiers:
\begin{itemize}
\item \textbf{FLUX-Schnell} and \textbf{FLUX-Dev}: Open-source diffusion models 
with strong baseline quality~\cite{flux2025}.
\item \textbf{Qwen-Image}: Multimodal autoregressive model with native 
text-to-image generation~\cite{wu2025qwen}.
\item \textbf{Z-Image-Turbo}: Novelly published, open-source diffusion model 
with fast inference speed and one of the best baseline quality ~\cite{cai2025zimage}.
\item \textbf{FLUX-2 Pro}: Commercial API with enhanced text rendering and 
compositional capabilities.
All models are accessed via their respective APIs with default parameters 
unless otherwise specified in ablation studies.
\end{itemize}
\paragraph{Image Editing Model:}
For CRAFT image editing experiments, we use \textbf{Gemini 2.5 Flash} by Google 
for its strong instruction-following in image manipulation tasks and efficient 
inference.

%------------------------------------------------------------------------------
% Hyperparameters and Configuration
%------------------------------------------------------------------------------
\subsection{Hyperparameters and Configuration}
\label{sec:hyperparameters}

\paragraph{Temperature and Sampling:}
All LLM and VLM calls use \texttt{temperature=0.0} for deterministic, 
reproducible outputs. This ensures consistency across runs and eliminates 
sampling variance in evaluation metrics.

\paragraph{Generation Parameters:}
For reproducibility, we use fixed seeds per prompt across all baseline and 
CRAFT configurations. T2I model parameters (guidance scale, inference steps) 
follow default API settings for each backbone to ensure fair comparison with 
typical user experiences. Questions used for DSG and DVQ metrics are generated 
using once for baseline prompt and use for evaluate in both baseline and CRAFT.

%------------------------------------------------------------------------------
% System Prompts
%------------------------------------------------------------------------------
\subsection{System Prompts}
\label{sec:system_prompts_overview}

Each component of the CRAFT pipeline uses a carefully designed system prompt 
to guide the LLM or VLM behavior:

\begin{itemize}
\item \textbf{Initial Prompt Rewriting}: Transforms user prompts into 
T2I-optimized format while preserving intent
(Appendix~\ref{app:prompt_rewrite_init}).
\item \textbf{Verify and Self Correct}: Verify whether the original prompt satisfies the constraints specified 
by the user and self-correct the prompt if necessary 
(Appendix~\ref{app:prompt_verify_and_self_correct}).
\item \textbf{DSG Question Generation}: Constructs visual questions and 
dependency graphs from prompts, as it was implemented in \cite{cho2024davidsonian}.
\item \textbf{VQA Evaluation}: Performs binary YES/NO visual question answering 
with rationales (Appendix~\ref{app:prompt_vqa}).
\item \textbf{DVQ Rationalization}: Rationalize the DVQ answers with answer "NO"
to the image (Appendix~\ref{app:prompt_dvq_rationalization}).
\item \textbf{Targeted Prompt Refinement}: Edits prompts based on DVQ feedback 
to address specific failures (Appendix~\ref{app:prompt_refinement}).
\item \textbf{Side-by-Side Judge}: Compares two images to select the better 
match for a given prompt (Appendix~\ref{app:prompt_judge}).
\item \textbf{Image Editing}: Applies targeted modifications to images based 
on structured instructions (Appendix~\ref{app:prompt_editing}).
\end{itemize}

Full prompt templates are provided in Appendix~\ref{app:system_prompts} to 
ensure reproducibility.

%------------------------------------------------------------------------------
% Computational Cost
%------------------------------------------------------------------------------
\subsection{Computational Cost}
\label{sec:computational_cost}

The additional inference-time cost introduced by CRAFT reasoning is negligible 
compared to T2I generation costs. For a typical $2048\times2048$ image with 
2 iterations:
\begin{itemize}
\item \textbf{VLM cost per iteration}: $\approx$\$0.0012 (vision tokens + 
text output for $2048\times2048$ images)
\item \textbf{LLM cost per iteration}: $\approx$\$0.00001 (prompt rewriting)
\item \textbf{Total CRAFT overhead}: $\approx$\$0.0024 for 2 iterations
\end{itemize}
This represents approximately 0.6\% of the cost of generating a single 
$2048\times2048$ image with premium T2I models like Hunyuan Image 3.0 
(\$0.42 per image). The cost analysis is detailed further in 
Section~\ref{sec:cost}.

\paragraph{Latency.}
A single CRAFT iteration (generation or editing) takes $\approx 30$ seconds on
average in our setup; three iterations typically yield strong results. In
contrast, a single Hunyuan Image 3.0 generation averages $\approx 60$ seconds
while underperforming CRAFT-equipped mid-cost models on compositional quality.

%==============================================================================
% Section: Metrics
%==============================================================================

\section{Metrics}
\label{sec:metrics}

%------------------------------------------------------------------------------
% VQA
%------------------------------------------------------------------------------
\subsection{VQA (DVQ / Visual Question Answering)}
\label{sec:vqa}

The VQA-based metric evaluates an image by asking a multimodal LLM a set of 
deterministic, unambiguous visual questions derived from the prompt. The model 
answers each question with a strict ``yes'' or ``no.'' The resulting yes-rate 
serves as a direct diagnostic score indicating how well the image matches each 
explicit constraint in the prompt:
\begin{equation}
\text{score}_{\text{VQA}}(\mathcal{I}) = \frac{1}{|\mathcal{Q}|} 
\sum_{q \in \mathcal{Q}} \mathbb{1}[\text{answer}_q = \text{YES}]
\end{equation}
where $\mathcal{Q}$ is the set of visual questions and $\mathcal{I}$ is the 
generated image.

%------------------------------------------------------------------------------
% DSG
%------------------------------------------------------------------------------
\subsection{DSG (Davidsonian Scene Graph)}
\label{sec:dsg}

DSG (Davidsonian Scene Graph) measures prompt--image consistency by converting 
a prompt into a set of structured, binary visual questions. These questions are 
arranged as a dependency graph $\mathcal{G}$, where a question is evaluated only 
if its parent conditions hold (for example, an attribute such as color is checked 
only after the corresponding object is confirmed to exist). The DSG score is 
computed as:
\begin{equation}
\text{score}_{\text{DSG}}(\mathcal{I}) = \frac{\sum_{q \in \mathcal{Q}_{\text{valid}}} 
\mathbb{1}[\text{answer}_q = \text{YES}]}{|\mathcal{Q}_{\text{valid}}|}
\end{equation}
where $\mathcal{Q}_{\text{valid}}$ is the set of questions whose parent conditions 
are satisfied, yielding a compositional, hierarchy-aware estimate of image 
faithfulness.

%------------------------------------------------------------------------------
% Auto Side-by-Side
%------------------------------------------------------------------------------
\subsection{Auto Side-by-Side}
\label{sec:auto_sxs}

Auto Side-by-Side is an automatic preference metric inspired by human A/B 
evaluations. A multimodal LLM is shown two images $\mathcal{I}_{\text{baseline}}$ 
and $\mathcal{I}_{\text{CRAFT}}$ along with the original prompt $\mathcal{P}$ 
and asked to select which image better satisfies the instruction. The win-rate 
is computed as:
\begin{equation}
\text{win-rate}_{\text{CRAFT}} = \frac{1}{N} \sum_{i=1}^{N} 
\mathbb{1}[\text{Judge}(\mathcal{I}_i^{\text{CRAFT}}, \mathcal{I}_i^{\text{baseline}}, 
\mathcal{P}_i) = \mathcal{I}_i^{\text{CRAFT}}]
\end{equation}
where $N$ is the total number of test prompts. This produces a binary preference 
score that correlates with human judgments while remaining model-agnostic.

%==============================================================================
% Section: Results
%==============================================================================

\section{Results}
\label{sec:results}

%==============================================================================
% Subsection: DSG1K Results
%==============================================================================

\subsection{Results on DSG1K Dataset}
\label{sec:results_dsg1k}

Table~\ref{tab:dsg1k} demonstrates that CRAFT yields consistent and substantial 
improvements across all evaluated model backbones on the DSG1K benchmark. For 
every configuration -- ranging from lighter diffusion models such as 
FLUX-Schnell and FLUX-Dev to more advanced systems like Qwen-Image, 
Z-Image-Turbo, and FLUX-2 Pro -- the iterative refinement pipeline increases 
both VQA and DSG scores and leads to pronounced gains in preference-based 
metrics.

\begin{table}[htbp]
\centering
\caption{Results on DSG1K~\cite{cho2024davidsonian} dataset.}
\label{tab:dsg1k}
\small
\begin{tabular}{@{}lcccccccccc@{}}
\toprule
\textbf{Metric} & \textbf{B1} & \textbf{C1} & \textbf{B2} & \textbf{C2} & 
\textbf{B3} & \textbf{C3} & \textbf{B4} & \textbf{C4} & \textbf{B5} & 
\textbf{C5} \\
\midrule
VQA & 0.78 & \textbf{0.86} & 0.77 & \textbf{0.87} & 0.864 & \textbf{0.946} & 
0.875 & \textbf{0.915} & 0.88 & \textbf{0.925} \\
DSG & 0.786 & \textbf{0.857} & 0.765 & \textbf{0.841} & 0.858 & 
\textbf{0.932} & 0.864 & \textbf{0.9} & 0.87 & \textbf{0.91} \\
Auto SxS & 0.21 & \textbf{0.744} & 0.34 & \textbf{0.65} & 0.36 & 
\textbf{0.59} & 0.27 & \textbf{0.67} & 0.4 & \textbf{0.52} \\
Auto SxS Adv & 0.26 & \textbf{0.74} & 0.288 & \textbf{0.671} & 0.33 & 
\textbf{0.58} & 0.328 & \textbf{0.63} & 0.47 & \textbf{0.5} \\
\bottomrule
\end{tabular}
\end{table}

\subsection{Comparison with Prompt Optimization Methods}
\label{sec:maestro_comparison}

Table~\ref{tab:maestro_comparison} compares CRAFT with prompt-optimization baselines
reported in Maestro~\cite{liu2025maestro}. We report only DSGScore and omit ranks.
Note that our CRAFT results use a GPT-based VLM judge (gpt-5-nano-2025-08-07),
whereas Maestro reports scores with Gemini 2.0 Flash, so absolute values are not
directly comparable across papers.

\begin{table}[htbp]
\centering
\caption{Comparison with prompt-optimization methods from Maestro. We report DSGScore
only; higher is better.}
\label{tab:maestro_comparison}
\small
\begin{tabular}{@{}lcc@{}}
\toprule
\textbf{Method} & \textbf{P2-hard DSGScore} $\uparrow$ & \textbf{DSG-1K DSGScore} $\uparrow$ \\
\midrule
Original & 0.826 & 0.772 \\
Rewrite & 0.855 & 0.815 \\
Promptist~\cite{hao2022optimizing} & 0.873 & 0.849 \\
LM-BBO~\cite{liu2024language} & 0.859 & 0.806 \\
OPT2I~\cite{wang2024opt2i} & 0.900 & 0.838 \\
Maestro~\cite{liu2025maestro} & \textbf{0.92} & 0.882 \\
\textbf{CRAFT (Ours)} & 0.90 & \textbf{0.91} \\
\bottomrule
\end{tabular}
\end{table}

CRAFT achieves the best DSGScore on DSG-1K and is competitive on P2-hard, where
Maestro reports slightly higher scores. Overall, constraint-driven refinement
provides stronger compositional gains than generic prompt-rewrite or agentic
optimization methods.

\begin{figure}[htbp]
\centering
\includegraphics[width=0.8\textwidth]{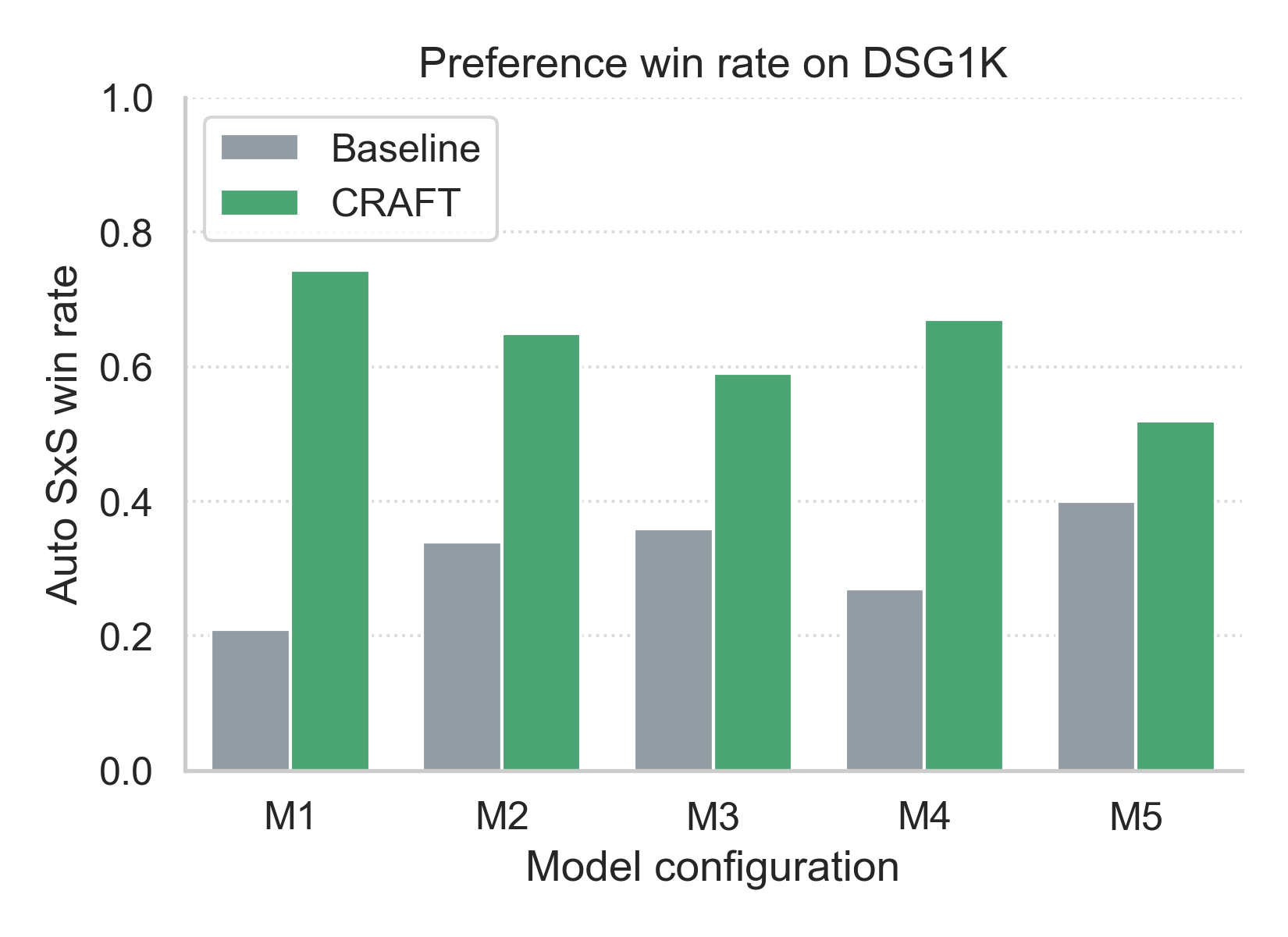}
\caption{Auto SxS win rates on DSG1K. CRAFT is preferred over baseline across 
all model configurations.}
\label{fig:dsg1k_wins}
\end{figure}

Notably, models with lower baseline compositional reliability benefit the most: 
for example, FLUX-Dev and FLUX-Schnell exhibit improvements of $8$--$10$ 
percentage points in VQA and DSG, with Auto SxS win-rates more than doubling 
(Figure~\ref{fig:dsg1k_wins}). At the same time, even stronger models such as 
Qwen-Image, Z-Image-Turbo, and FLUX-2 Pro continue to improve, indicating that 
the proposed approach scales with model capacity and is not limited to weaker 
backbones.

%==============================================================================
% Subsection: Parti-Prompt Results
%==============================================================================

\subsection{Results on Parti-Prompt Dataset}
\label{sec:results_parti}

Table~\ref{tab:parti} shows that the proposed CRAFT leads to consistent 
improvements across all evaluated model families on the Parti-Prompts dataset. 
For lighter T2I backbones such as FLUX-Dev and FLUX-Schnell, two reasoning 
iterations yield clear gains in both VQA and DSG scores, often closing much of 
the gap to stronger baselines. Higher-capacity models such as Qwen-Image and 
FLUX-2-Pro also benefit from iterative refinement, though their improvements are 
smaller in absolute magnitude due to already strong baseline performance.

\begin{table}[htbp]
\centering
\caption{Results on Parti-Prompt dataset.}
\label{tab:parti}
\small
\begin{tabular}{@{}lcccccccccc@{}}
\toprule
\textbf{Metric} & \textbf{B1} & \textbf{C1} & \textbf{B2} & \textbf{C2} & 
\textbf{B3} & \textbf{C3} & \textbf{B4} & \textbf{C4} & \textbf{B5} & 
\textbf{C5} \\
\midrule
VQA & 0.79 & \textbf{0.88} & 0.822 & \textbf{0.911} & 0.89 & \textbf{0.94} & 
0.9 & \textbf{0.94} & 0.90 & \textbf{0.92} \\
DSG & 0.743 & \textbf{0.857} & 0.815 & \textbf{0.895} & 0.896 & 
\textbf{0.939} & 0.89 & \textbf{0.92} & 0.87 & \textbf{0.90} \\
Auto SxS & 0.19 & \textbf{0.756} & 0.382 & \textbf{0.618} & 0.445 & 
\textbf{0.555} & 0.268 & \textbf{0.679} & 0.42 & \textbf{0.47} \\
Auto SxS Adv & 0.252 & \textbf{0.748} & 0.309 & \textbf{0.628} & 0.41 & 
\textbf{0.53} & 0.328 & \textbf{0.668} & 0.47 & \textbf{0.52} \\
\bottomrule
\end{tabular}
\end{table}

\begin{figure}[htbp]
\centering
\includegraphics[width=0.8\textwidth]{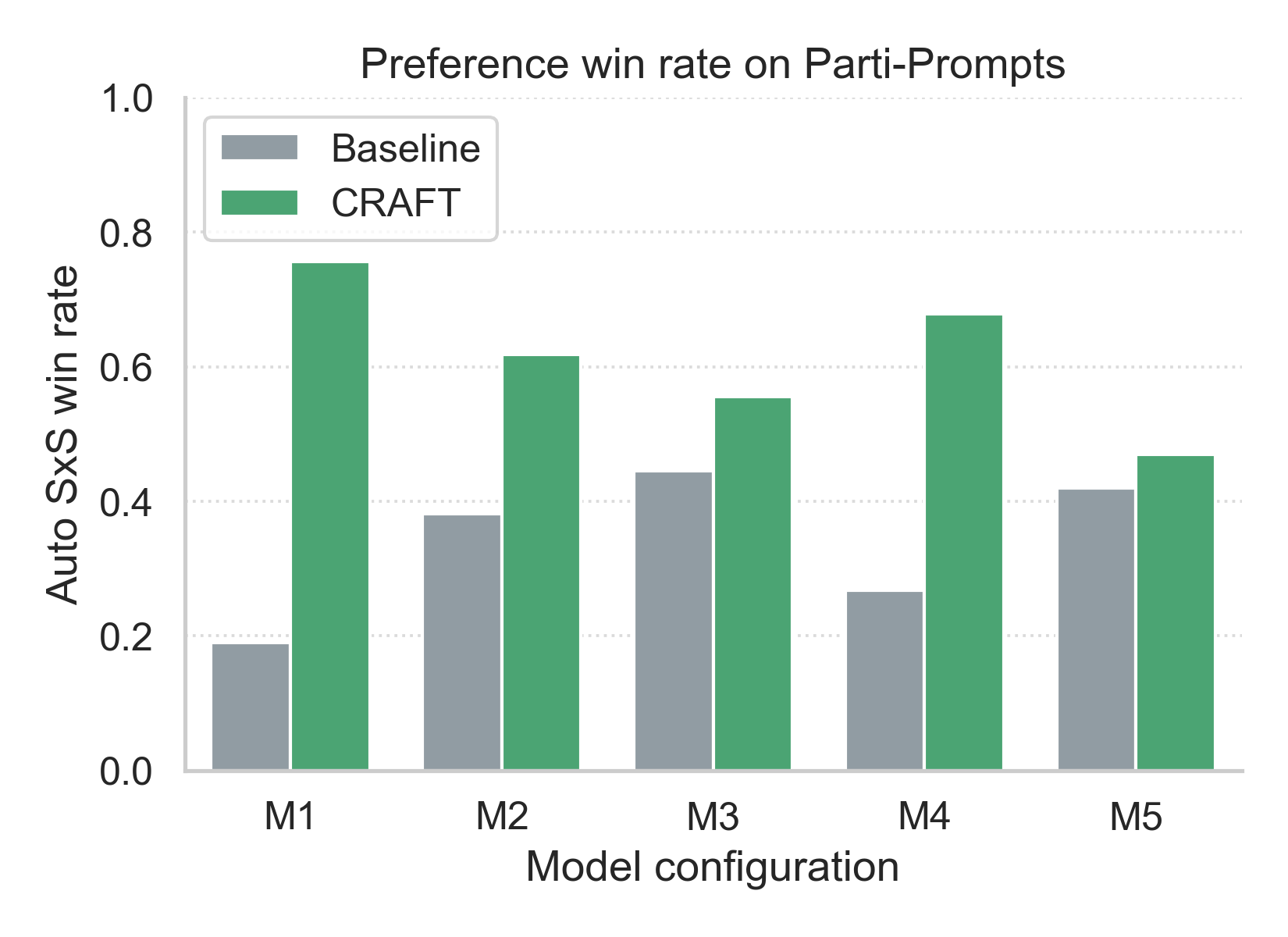}
\caption{Auto SxS win rates on Parti-Prompt. CRAFT is preferred over baseline 
across all model configurations.}
\label{fig:parti_wins}
\end{figure}

Preference-based metrics (Auto SxS and Auto SxS advantage) indicate that CRAFT 
is systematically preferred over baseline generations for most models 
(Figure~\ref{fig:parti_wins}), with particularly large shifts for weaker 
backbones. Qualitative comparisons further align with these quantitative trends: 
across all architectures, CRAFT reduces omissions, improves attribute 
consistency, and produces more coherent text rendering and layout for text-heavy 
prompts.

%==============================================================================
% Subsection: Ablation - Effect of Number of Iterations
%==============================================================================

\subsection{Ablation: Effect of Number of Iterations}
\label{sec:ablation_iters}

\begin{figure}[t]
\centering
\includegraphics[width=0.7\textwidth]{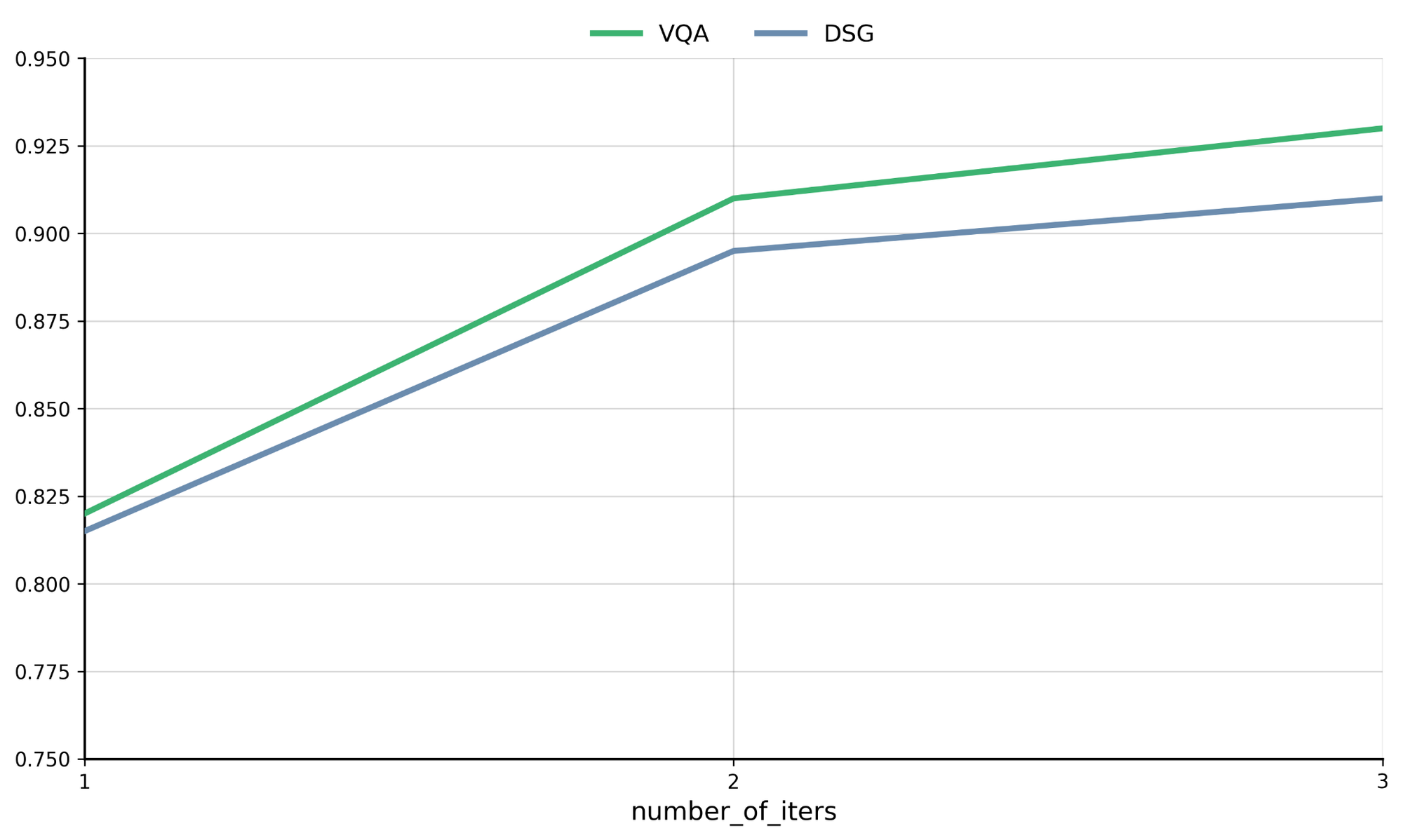}
\caption{VQA and DSG scores as a function of the number of iterations for CRAFT 
with FLUX-Schnell.}
\label{fig:iterations}
\end{figure}

Figure~\ref{fig:iterations} demonstrates that the introduction of even a single 
reasoning iteration yields a substantial improvement in both VQA and DSG scores 
relative to the one-shot FLUX baseline. Additional iterations continue to 
provide incremental gains, though with diminishing returns, and importantly do 
not degrade performance. This behavior suggests that the ``thinking'' loop is 
most effective in its first pass -- correcting major prompt--image 
inconsistencies -- while subsequent iterations refine remaining details without 
destabilizing generation quality.

%==============================================================================
% Subsection: Text Rendering Quality
%==============================================================================

\subsection{Text Rendering Quality: Normalized Edit Distance}
\label{sec:text_quality}

We evaluate text rendering quality using Normalized Edit Distance (NED), 
computed as:
\begin{equation}
\text{NED}(x, y) = \frac{\text{Levenshtein}(x, y)}{\max(|x|, |y|)}
\end{equation}
where $\text{Levenshtein}(x, y)$ measures character-level edits (insertions, 
deletions, substitutions) and normalization ensures fair comparison across 
different phrase lengths. For each prompt, we extract expected text phrases 
using an LLM and perform OCR on generated images using a Vision-Language Model. 
We report Text Accuracy $= 1 - \text{NED}$ for intuitive interpretation and 
Exact Match Rate (percentage of phrases with $\text{NED} = 0$) as a strict 
quality measure.

Table~\ref{tab:text_quality} shows that CRAFT yields measurable improvements in 
text rendering performance for models with weaker baseline text fidelity. 
FLUX-Schnell exhibits the largest gains in both Text Accuracy and Exact Match 
rate, indicating that iterative constraint-based refinement reduces 
character-level deviations. FLUX-Dev shows moderate improvements, consistent 
with its baseline limitations. For Z-Image and FLUX-2 Pro, changes introduced by 
CRAFT are minor, reflecting their relatively stable baseline behavior and 
limited headroom for improvement in this task.

\begin{table}[htbp]
\centering
\caption{Text rendering quality evaluated using Normalized Edit Distance (NED).}
\label{tab:text_quality}
\begin{tabular}{@{}lcccc@{}}
\toprule
\textbf{Model} & \textbf{Baseline Acc.} & \textbf{CRAFT Acc.} & 
\textbf{Improvement} & \textbf{Exact Match Gain} \\
\midrule
FLUX-Schnell & 74.01\% & 81.17\% & \textbf{+7.16\%} & \textbf{+12.91\%} \\
FLUX-Dev & 64.01\% & 68.90\% & \textbf{+4.89\%} & \textbf{+6.00\%} \\
Z-Image & 91.00\% & 91.40\% & \textbf{+0.40\%} & \textbf{+1.10\%} \\
FLUX-2 Pro & 98.50\% & 98.80\% & \textbf{+0.30\%} & \textbf{+0.25\%} \\
\bottomrule
\end{tabular}
\end{table}

%==============================================================================
% Subsection: Human Evaluation
%==============================================================================

\subsection{Human Evaluation}
\label{sec:human_eval}

To complement the automatic preference metrics, we conducted an additional 
small-scale human evaluation on two model families: Z-Image-Turbo and FLUX-2 
Pro. Twenty evaluators (who work in the field of AI content generation) were 
shown two images (baseline and CRAFT mode) and asked to select the version that 
best satisfied the original prompt. Ties were allowed.

\begin{figure}[h!]
\centering
\includegraphics[width=0.7\textwidth]{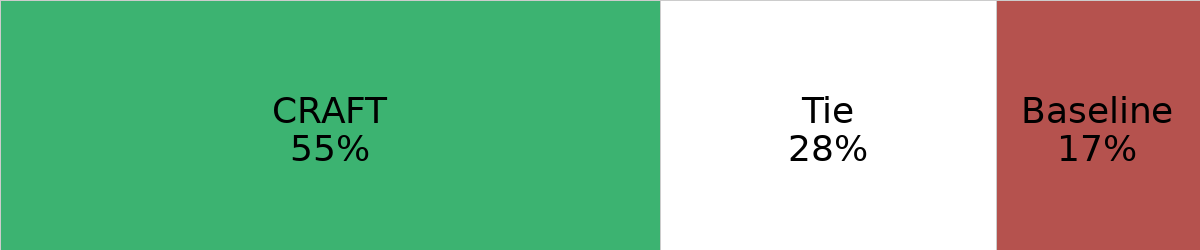}
\caption{Human evaluation results for FLUX-Schnell model.}
\label{fig:human_eval1}
\end{figure}

\begin{figure}[h!]
\centering
\includegraphics[width=0.7\textwidth]{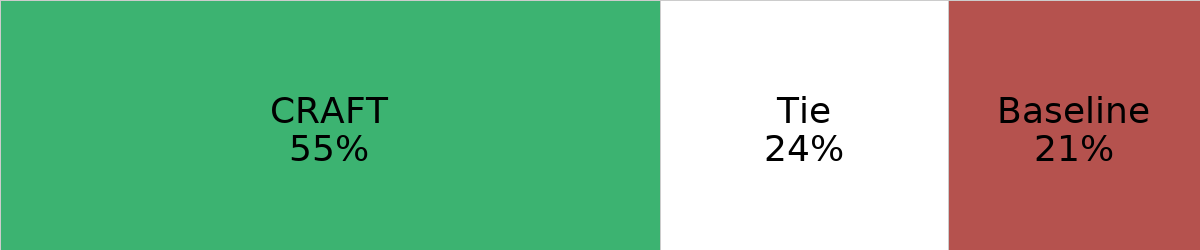}
\caption{Human evaluation results for FLUX-Dev model.}
\label{fig:human_eval2}
\end{figure}

\begin{figure}[h!]
\centering
\includegraphics[width=0.7\textwidth]{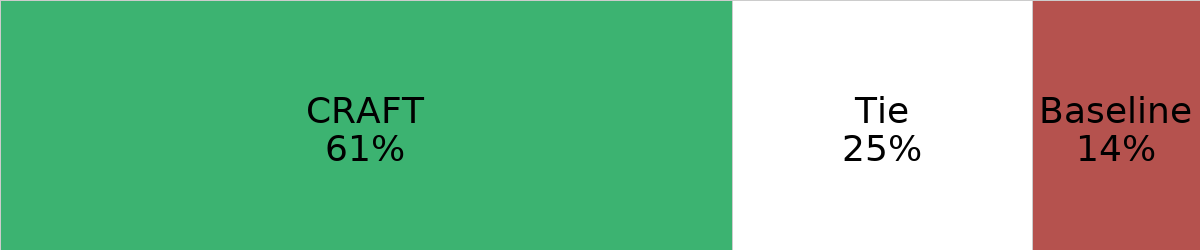}
\caption{Human evaluation results for Z-Image-Turbo model.}
\label{fig:human_eval3}
\end{figure}

\begin{figure}[h!]
\centering
\includegraphics[width=0.7\textwidth]{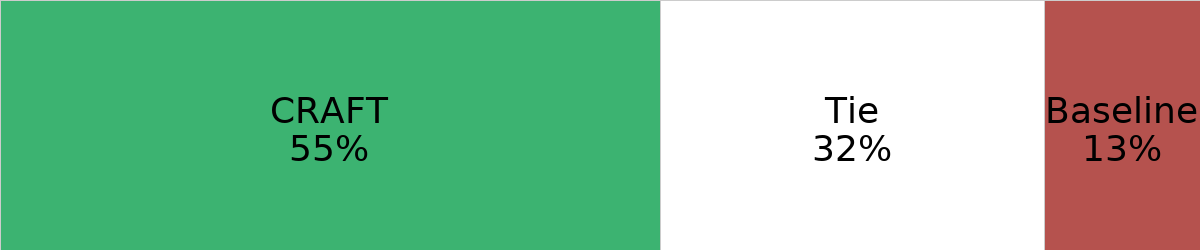}
\caption{Human evaluation results for FLUX-2 Pro model.}
\label{fig:human_eval4}
\end{figure}

Across both models, the results closely mirrored the automatic side-by-side 
metrics, consistently favoring the CRAFT outputs 
(Figures~\ref{fig:human_eval1}--\ref{fig:human_eval4}). These results indicate 
that the improvements observed under DVQ and side-by-side metrics are also 
reflected in human preference judgments.

%==============================================================================
% Subsection: Qualitative Results
%==============================================================================

\subsection{Qualitative Results}
\label{sec:qualitative}

We present qualitative comparisons across all evaluated models. Across 
backbones, reasoning-guided refinement yields improved attribute consistency, 
reduced omissions, more coherent spatial relations, and better handling of long 
prompts with multiple clauses or textual regions.

%------------------------------------------------------------------------------
% Qwen Image 2 iterations + Gemini Flash 2.5
%------------------------------------------------------------------------------

\subsubsection{Qwen Image 2 iterations + Gemini Flash 2.5}

\begin{figure}[!h]
\centering
\includegraphics[width=0.45\textwidth]{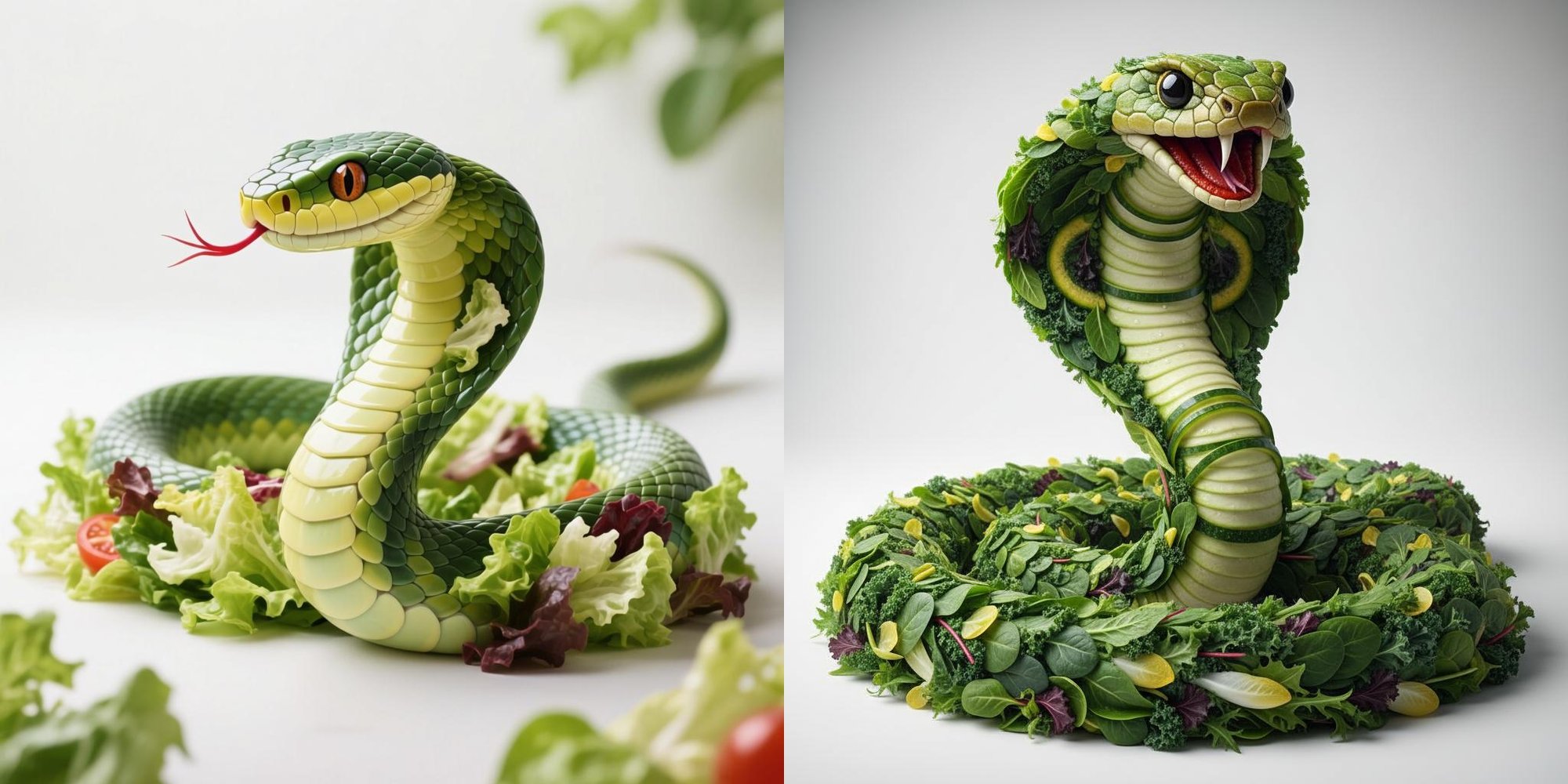}
\caption{Baseline vs CRAFT. Prompt: \textit{A giant cobra snake made from 
salad}.}
\label{fig:qual1}
\end{figure}

\begin{figure}[!h]
\centering
\includegraphics[width=0.45\textwidth]{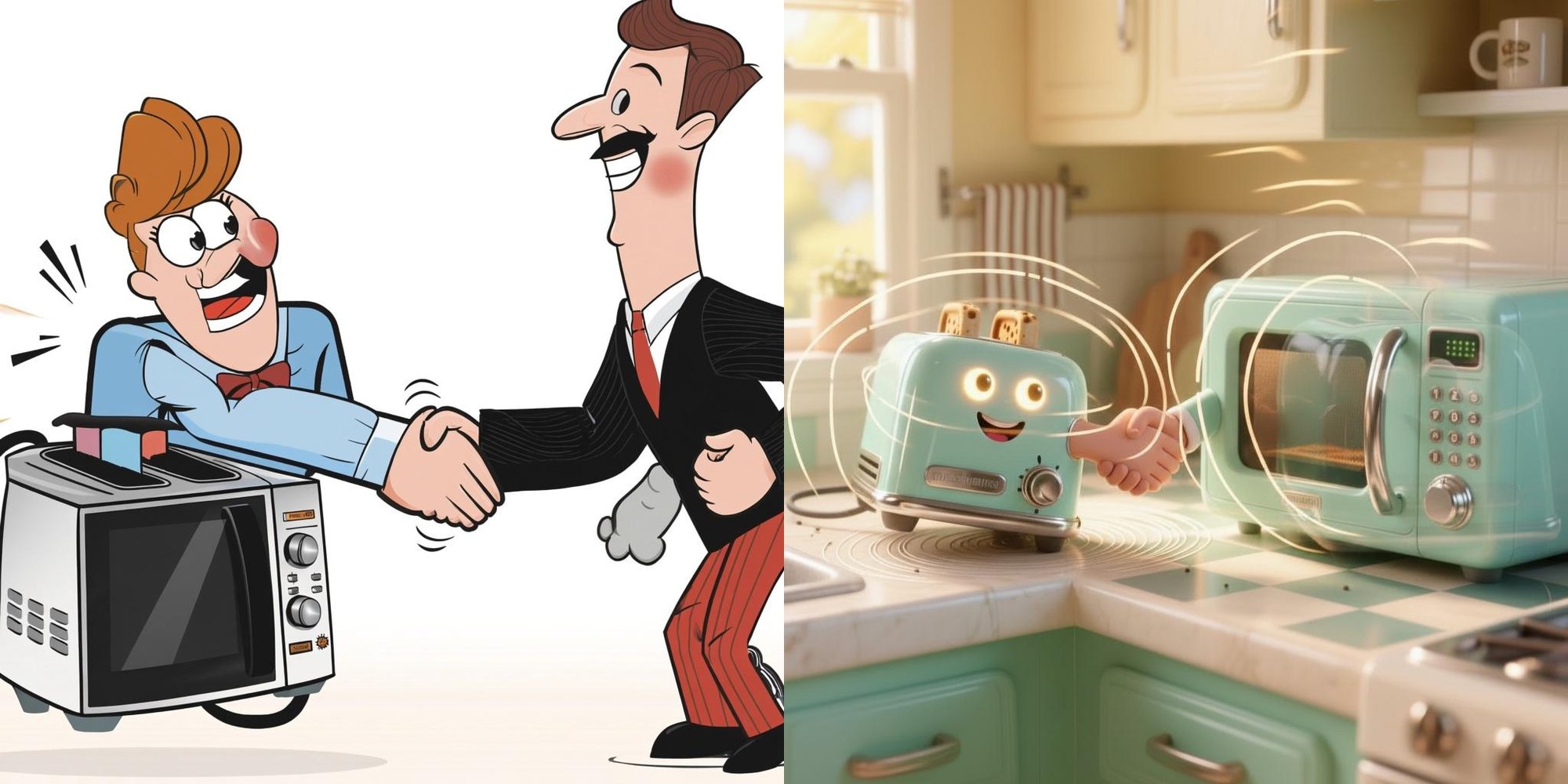}
\caption{Baseline vs CRAFT. Prompt: \textit{a toaster shaking hands with a 
microwave}.}
\label{fig:qual3}
\end{figure}

\begin{figure}[!h]
\centering
\includegraphics[width=0.45\textwidth]{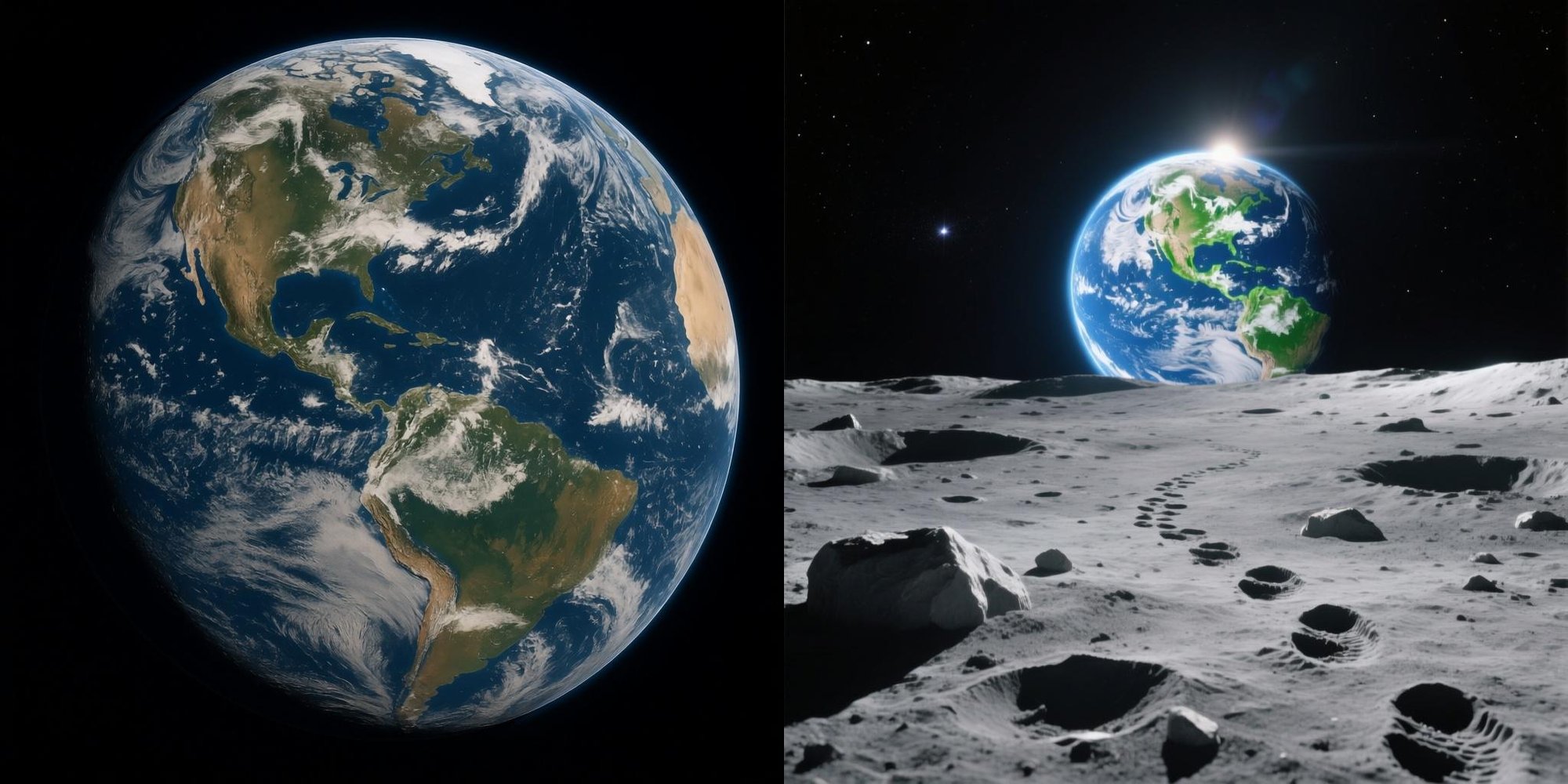}
\caption{Baseline vs CRAFT. Prompt: \textit{a view of the Earth from the moon}.}
\label{fig:qual6}
\end{figure}

% Text between figure groups
We also demonstrate results for long-form prompts and for prompts that include 
text elements rendered within the generated images (see 
Appendix~\ref{app:long_prompts} for full prompt listings).

\begin{figure}[!t]
\centering
\includegraphics[width=0.8\textwidth]{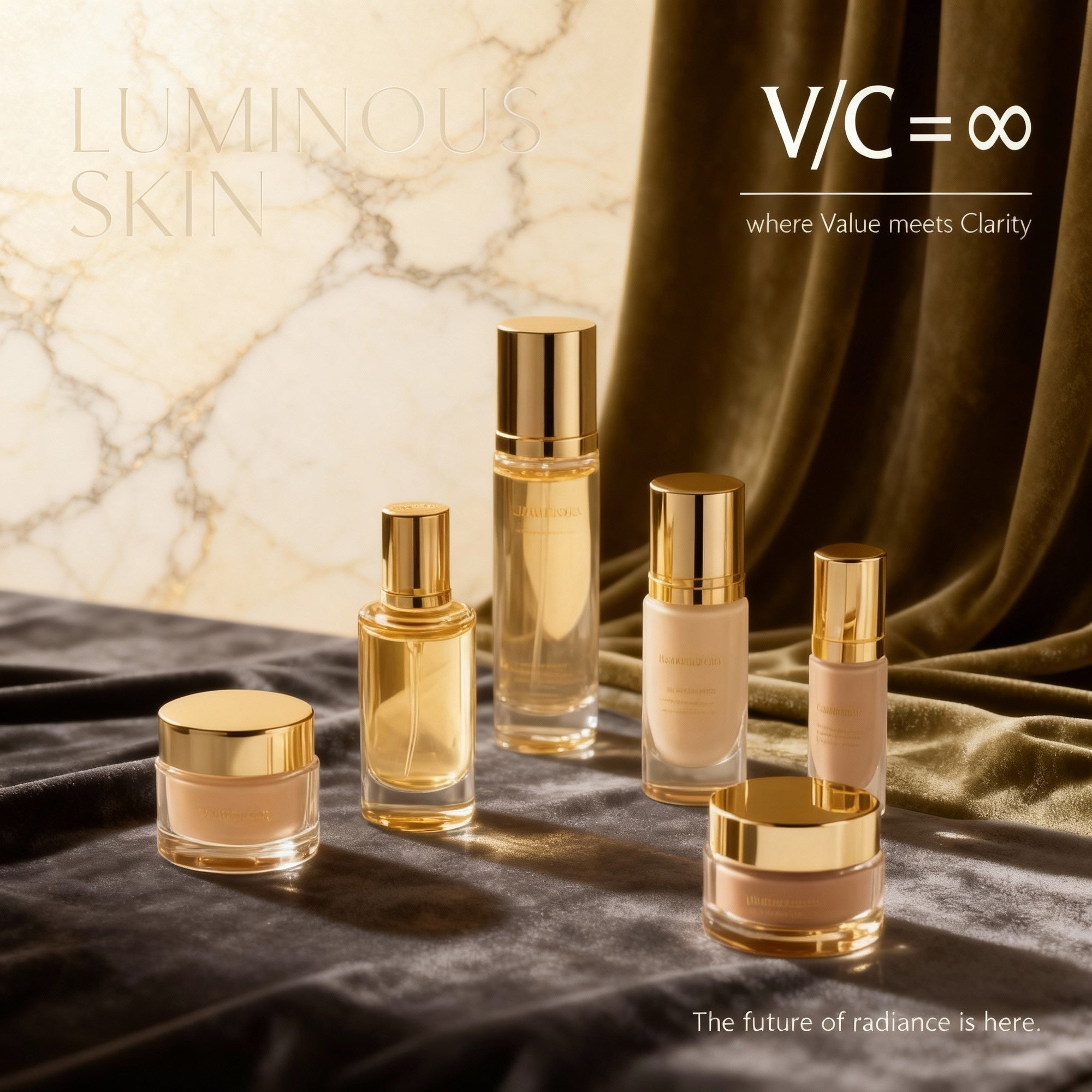}
\caption{CRAFT results for long\_prompt1.}
\label{fig:long1}
\end{figure}

\clearpage

%------------------------------------------------------------------------------
% FLUX-Dev 2 iterations
%------------------------------------------------------------------------------

\subsubsection{FLUX-Dev 2 iterations}

We also present a comparison between the baseline and the CRAFT (2 iterations) 
for FLUX-Dev, without using any image-editing models. All generations were 
produced with standard parameters. Even this lightweight setup -- simply 
correcting the prompt based on the VQ questions and applying the thinking 
procedure -- already leads to noticeable improvements in generation quality. 
Additional examples are provided in Appendix~\ref{app:additional_qualitative}.

\begin{figure}[!h]
\centering
\includegraphics[width=0.45\textwidth]{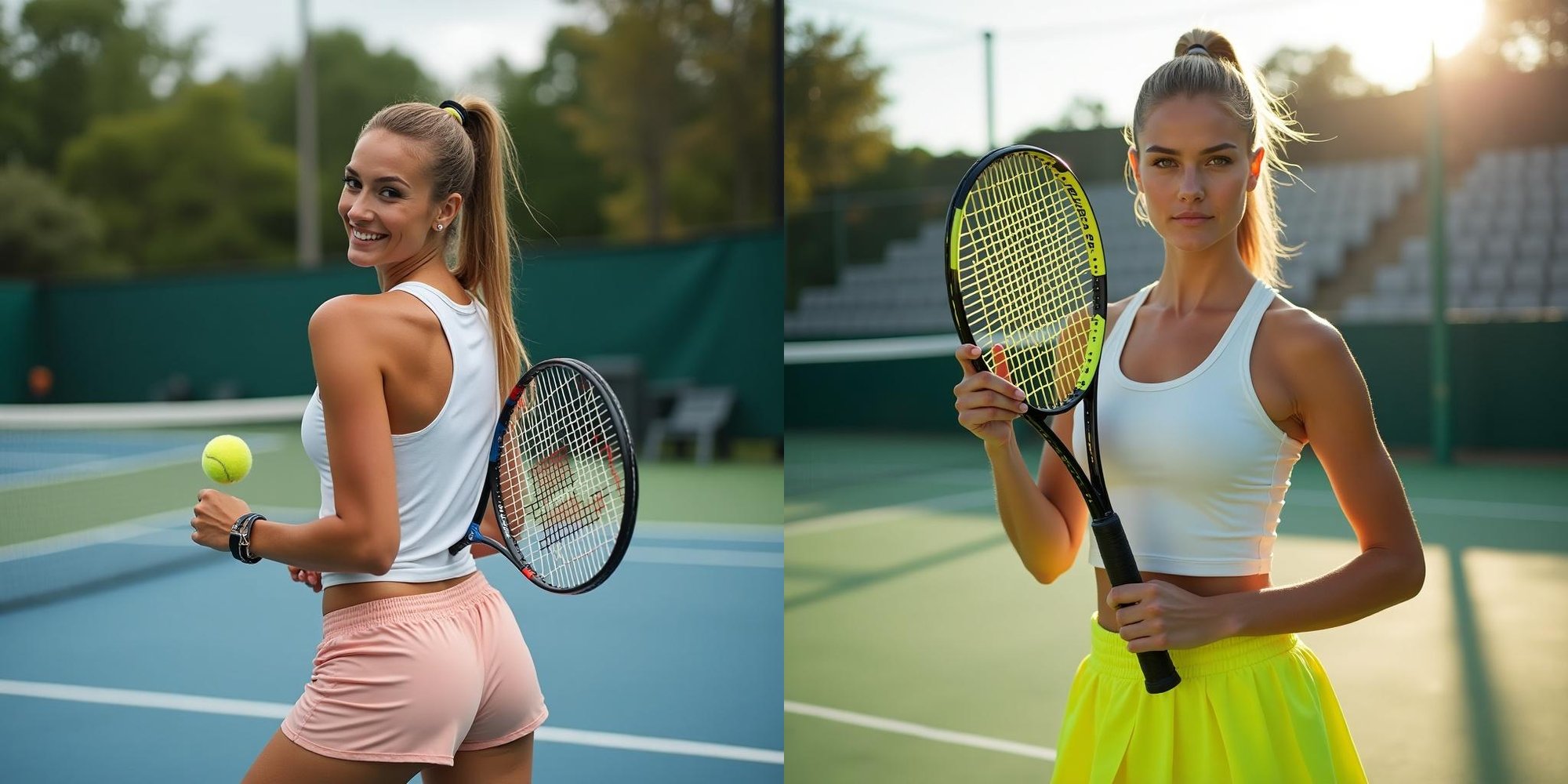}
\caption{Baseline vs CRAFT. Prompt: \textit{A woman holding a racquet on top of 
a tennis court}.}
\label{fig:fluxdev1}
\end{figure}

\begin{figure}[!h]
\centering
\includegraphics[width=0.45\textwidth]{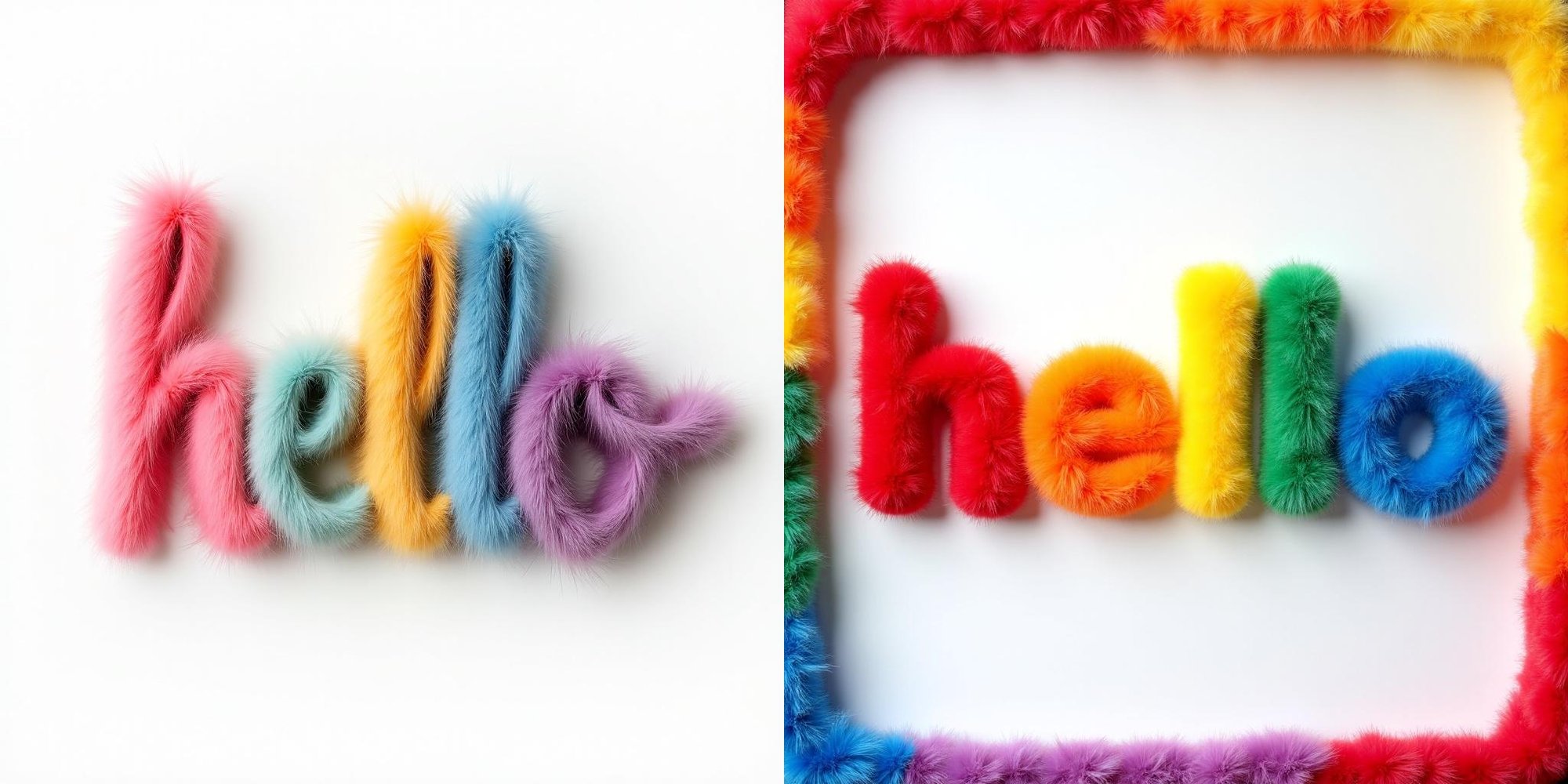}
\caption{Baseline vs CRAFT. Prompt: \textit{studio shot multicolored fur in the 
shape of word 'hello', in a furry frame, white background, centered}.}
\label{fig:fluxdev2}
\end{figure}

\begin{figure}[!h]
\centering
\includegraphics[width=0.45\textwidth]{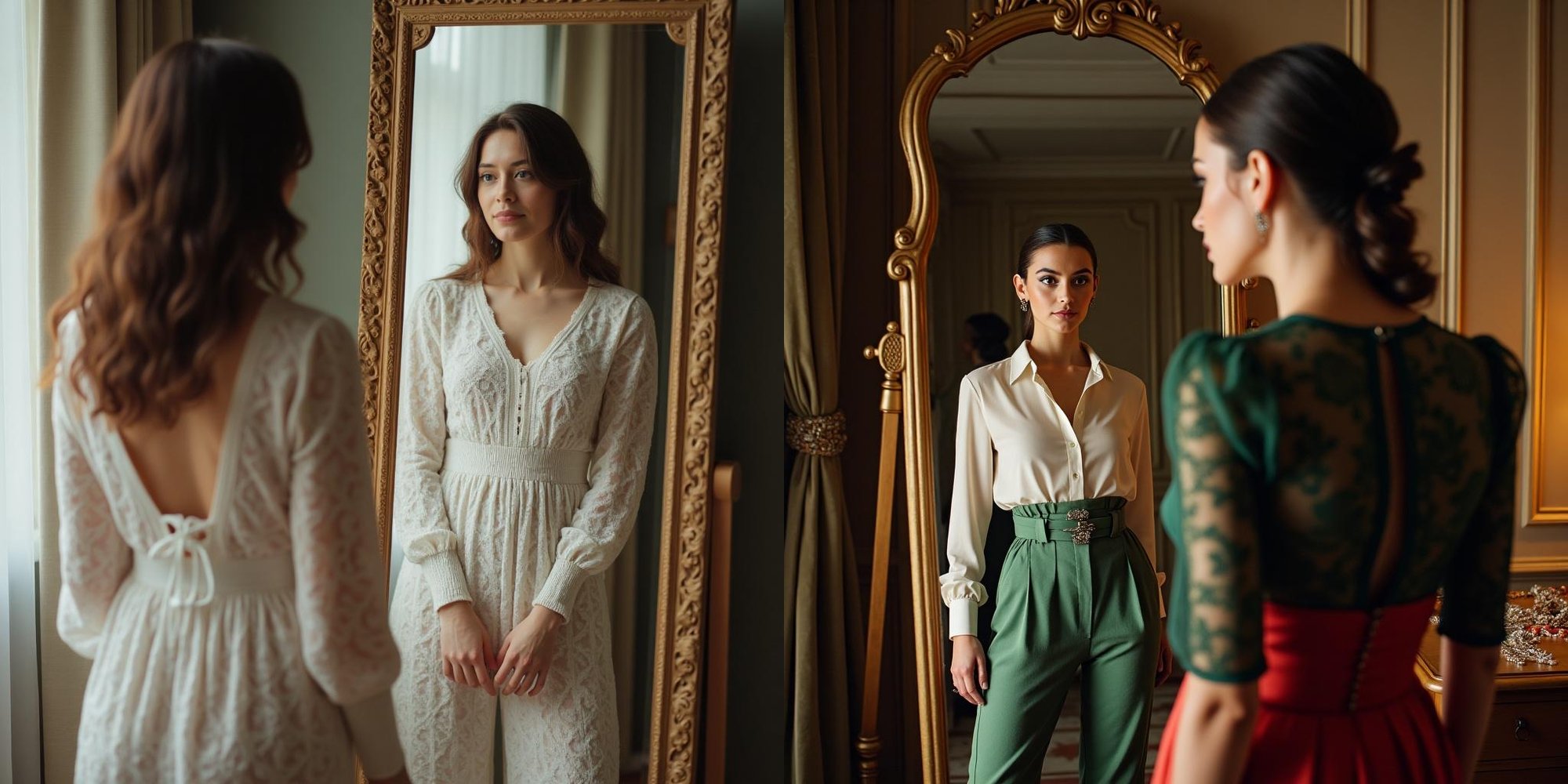}
\caption{Baseline vs CRAFT. Prompt: \textit{A woman's mirror reflection is 
wearing different clothes}.}
\label{fig:fluxdev3}
\end{figure}

\clearpage

%------------------------------------------------------------------------------
% Z-Image 2 iterations
%------------------------------------------------------------------------------

\subsubsection{Z-Image 2 iterations}

We additionally evaluate Z-Image under a two-iteration CRAFT, without any 
image-editing components. Additional examples are provided in 
Appendix~\ref{app:additional_qualitative}.

\begin{figure}[!h]
\centering
\includegraphics[width=0.45\textwidth]{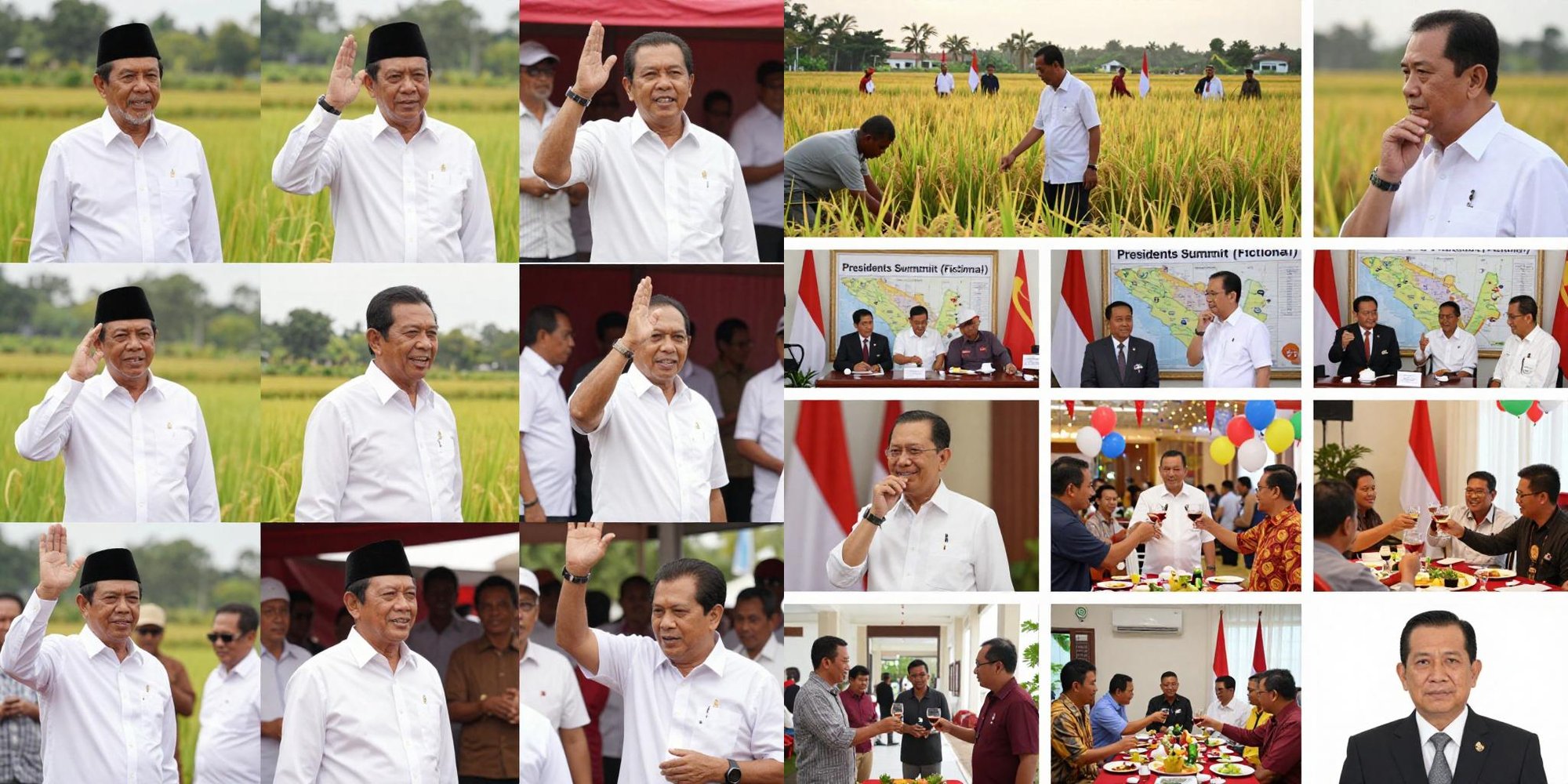}
\caption{Baseline vs CRAFT. Prompt: \textit{Generate a series of six candid, documentary-style photos of this Indonesian president in office, in the rice fields, and partying with other presidents.}}
\label{fig:zimage1}
\end{figure}

\begin{figure}[!h]
\centering
\includegraphics[width=0.45\textwidth]{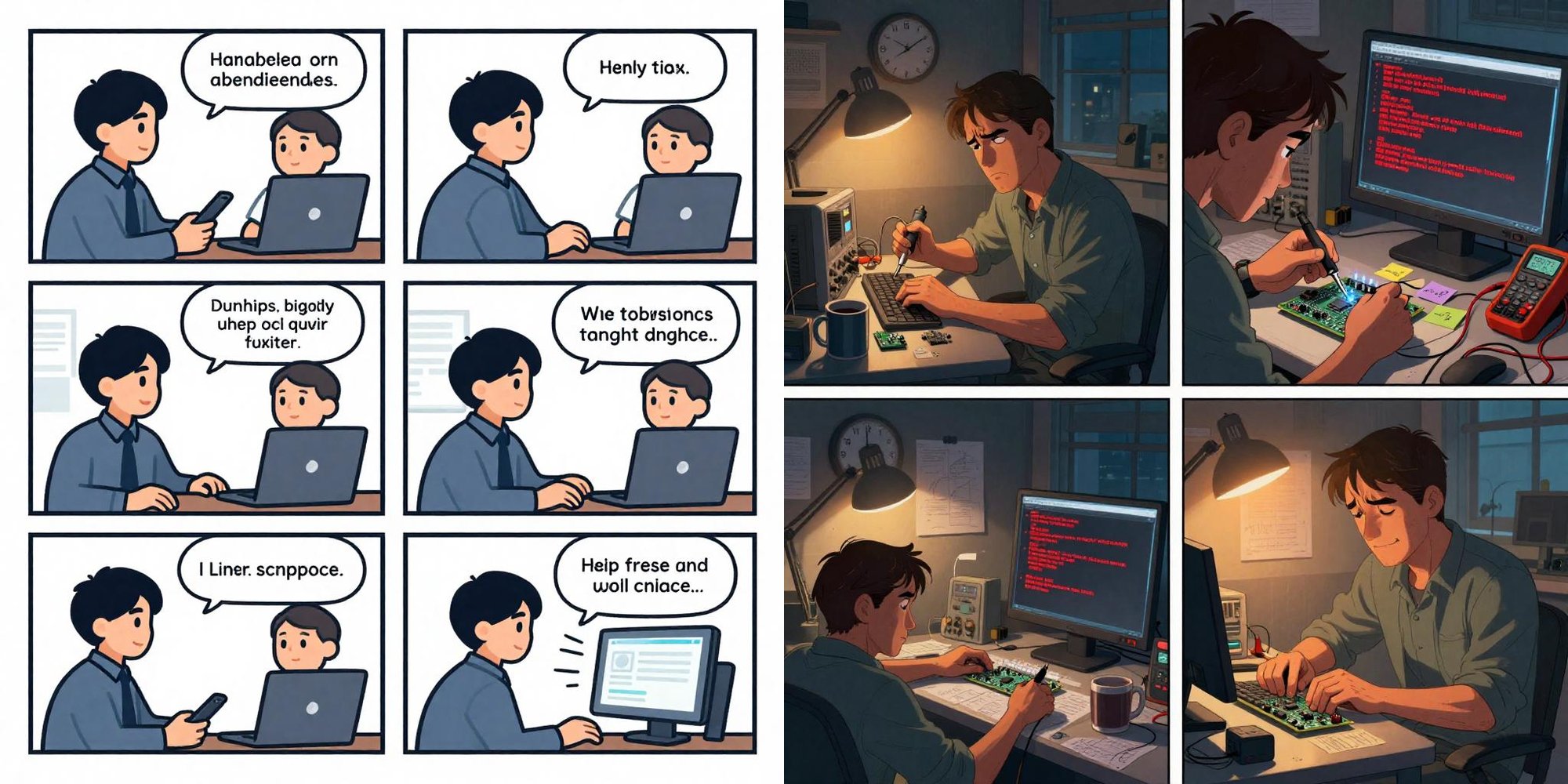}
\caption{Baseline vs CRAFT. Prompt: \textit{Generate a 4‑panel comic about the 
hardships of an embedded engineer}.}
\label{fig:zimage2}
\end{figure}

\clearpage

%------------------------------------------------------------------------------
% FLUX-2 Pro 2 iterations
%------------------------------------------------------------------------------

\subsubsection{FLUX-2 Pro 2 iterations}

We additionally evaluate FLUX2-Pro under a two-iteration CRAFT, without any 
image-editing components. Additional examples are provided in 
Appendix~\ref{app:additional_qualitative}.

\begin{figure}[!h]
\centering
\includegraphics[width=0.45\textwidth]{images/image30.jpg}
\caption{Baseline vs CRAFT. Prompt: \textit{Dream diary. A pink Kirby lying 
asleep on a star, blowing rainbow-colored bubbles from its mouth. Soft macaron 
color palette, cloud and candy stickers, glittery crayon-like details, dreamy 
and sweet}.}
\label{fig:flux2pro1}
\end{figure}

\begin{figure}[!h]
\centering
\includegraphics[width=0.45\textwidth]{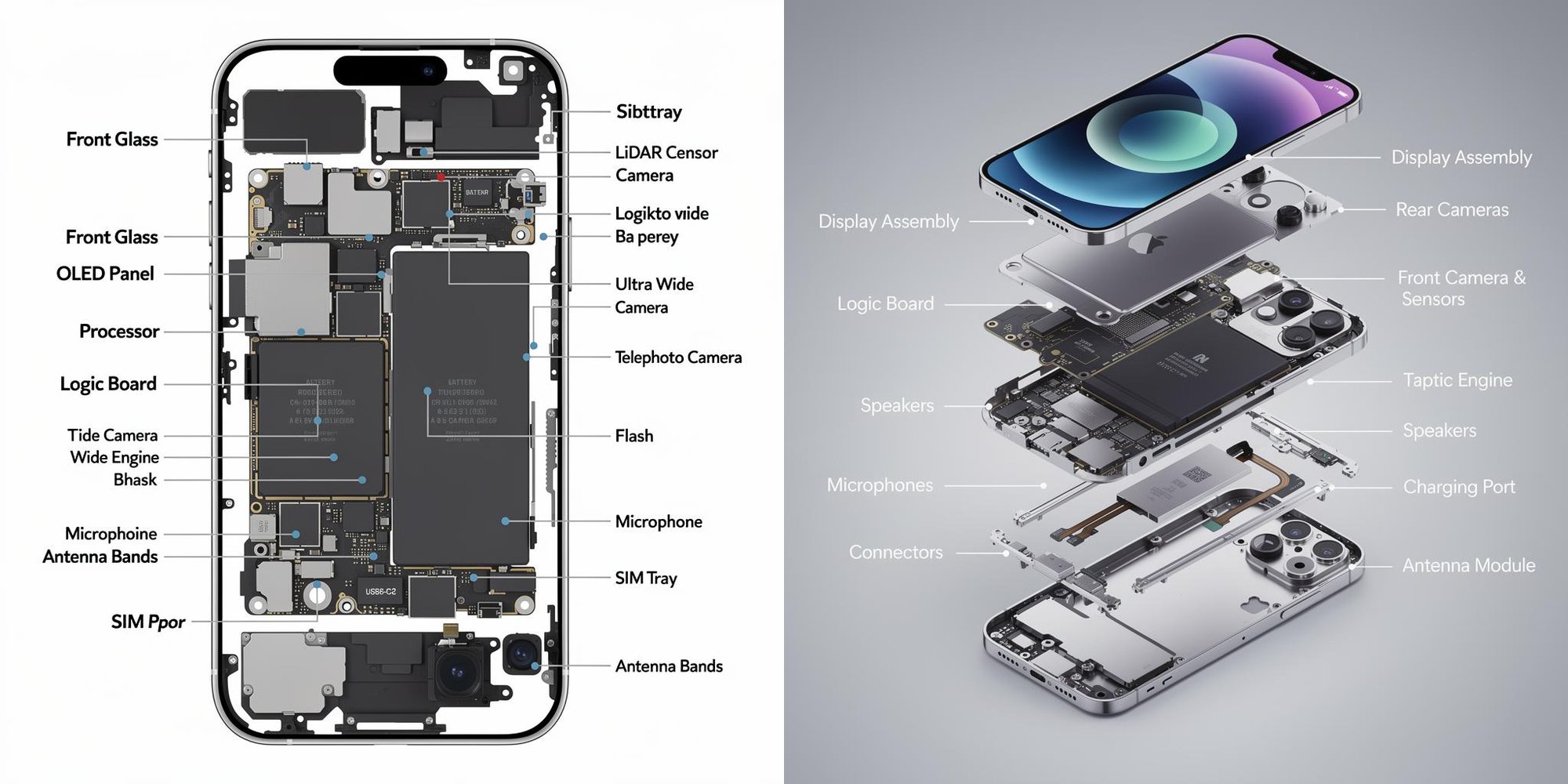}
\caption{Baseline vs CRAFT. Prompt: \textit{Create a technical exploded view 
diagram of an iPhone pro max in 1080x1080 dimensions. The device should be 
deconstructed, showing all individual components floating in space, separated to 
reveal the internal parts. Each major component is labeled}.}
\label{fig:flux2pro2}
\end{figure}

\clearpage

%------------------------------------------------------------------------------
% Abstract prompts filtering
%------------------------------------------------------------------------------

We observed that some prompts in the PartiPrompt dataset produce unreliable 
generations: several entries consist of only one or two highly abstract words, 
which lead to poor results in both the baseline and the CRAFT. Examples include 
prompts such as ``emotion'' or ``hate''. To address this, we filtered the 
PartiPrompt subset and retained only prompts containing more than three words.

\begin{figure}[!h]
\centering
\includegraphics[width=0.6\textwidth]{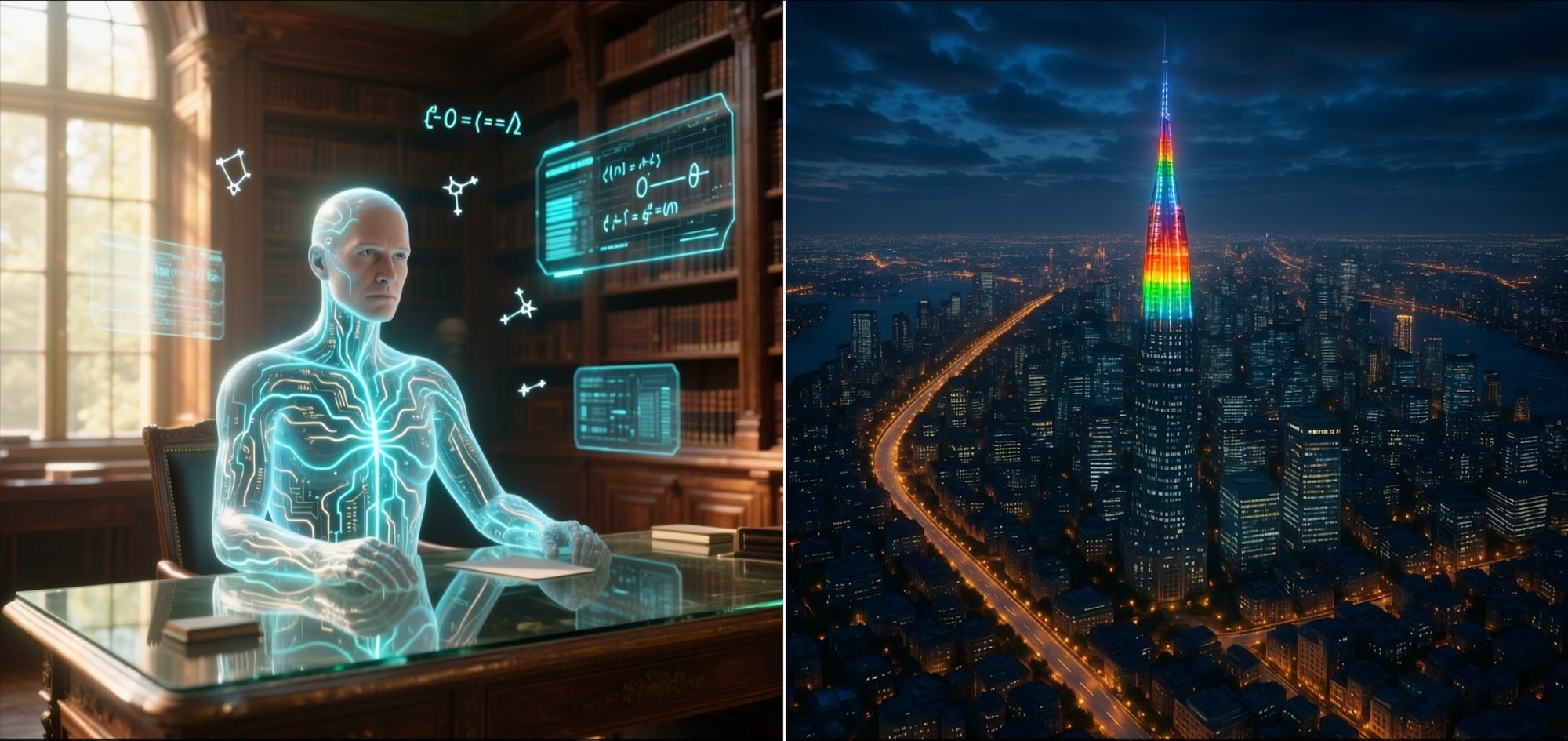}
\caption{CRAFT vs baseline for prompt: \textit{intelligence}.}
\label{fig:abstract1}
\end{figure}

\begin{figure}[!h]
\centering
\includegraphics[width=0.6\textwidth]{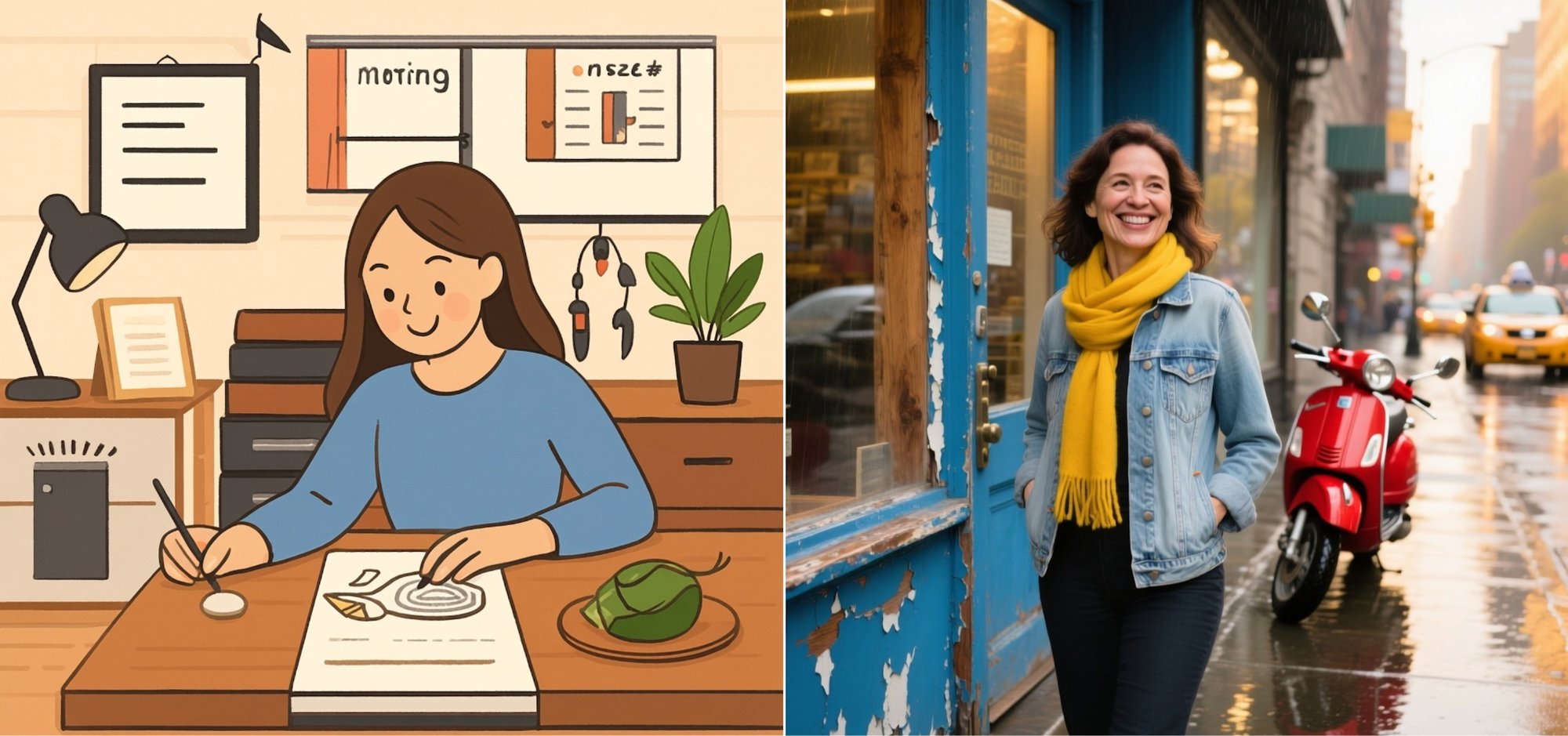}
\caption{Baseline vs CRAFT for prompt: \textit{commonsense}.}
\label{fig:abstract2}
\end{figure}

%------------------------------------------------------------------------------
% GenEval detector issues
%------------------------------------------------------------------------------

We conducted evaluations using GenEval, but the results were not representative, 
as the available detectors performed poorly even on clear, high-quality cases. 
The detectors used were those provided in MMCV 1.7.1, including QueryInst R101, 
Mask2Former, and YOLOX-L.

\begin{figure}[!h]
\centering
\includegraphics[width=0.6\textwidth]{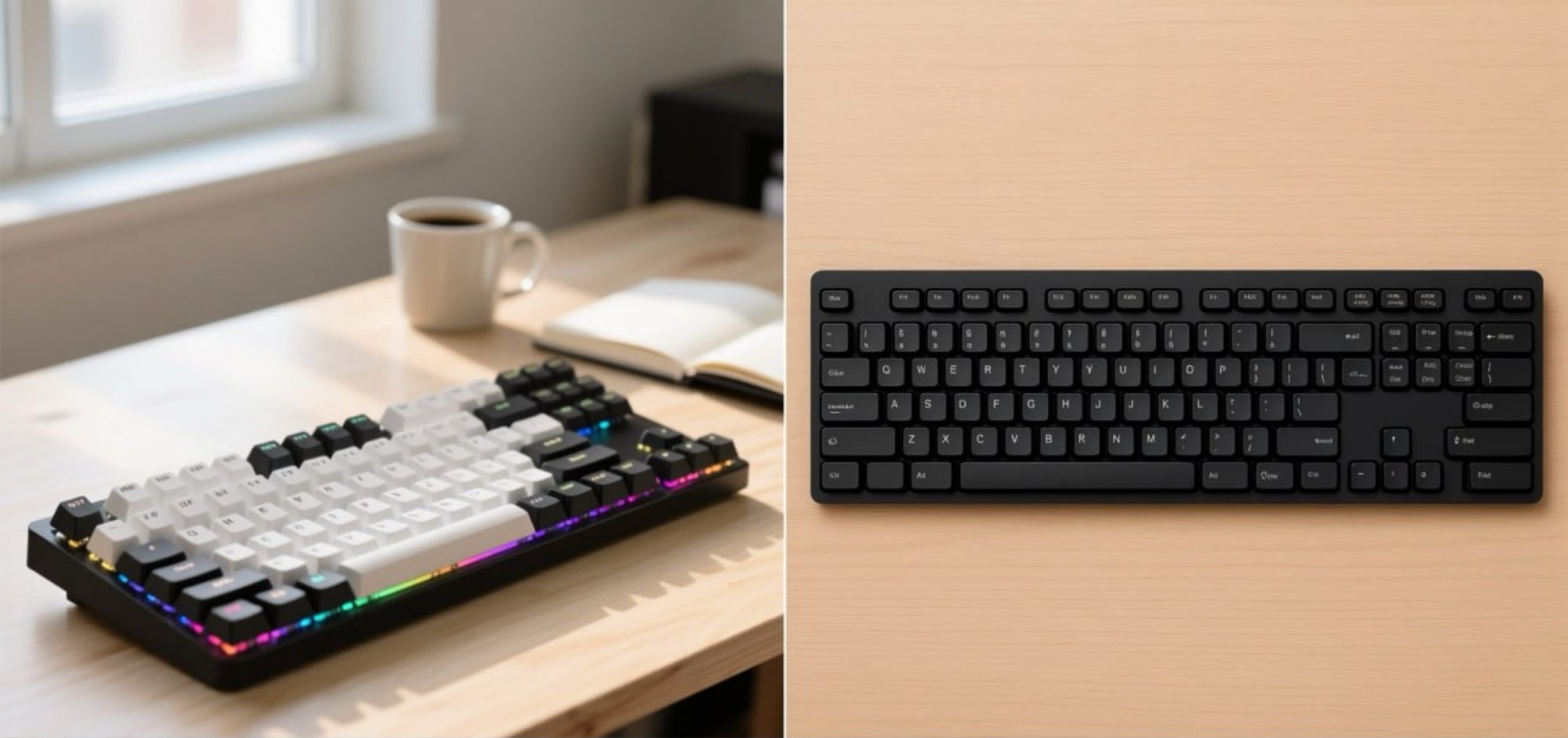}
\caption{CRAFT versus baseline for prompt: \textit{a photo of a keyboard}.}
\label{fig:geneval}
\end{figure}

%==============================================================================
% Section: Cost and Quality Comparison
%==============================================================================

\section{Cost and Quality Comparison}
\label{sec:cost}

To complement the compositional and preference-based evaluations, we analyze 
the cost--quality trade-off of applying CRAFT to lightweight T2I generators. 
Table~\ref{tab:cost_quality_2048} presents a detailed breakdown of generation 
costs and quality gains at a fixed resolution of $2048\times2048$ pixels 
($\approx 4.19$ megapixels) across five backbone models.

\subsection{Cost Model}

For a backbone with price $c$ dollars per megapixel (MP), the cost of generating 
a single image at resolution $r$ MP is $c \cdot r$. CRAFT with $T$ iterations 
requires $T$ generation calls plus multimodal reasoning overhead. For 
$T{=}2$ iterations at $r{=}4.19$ MP, the total cost becomes:
\begin{equation}
\text{Cost}_{\text{CRAFT}} = 2 \cdot c \cdot r + \text{Cost}_{\text{reasoning}}
\end{equation}
where $\text{Cost}_{\text{reasoning}} \approx \$0.0024$ (two VLM evaluation 
cycles using GPT-5-nano for $2048\times2048$ images).

\subsection{Empirical Cost--Quality Analysis}

Table~\ref{tab:cost_quality_2048} reveals four key patterns:

\paragraph{Mid-cost generators achieve premium quality below premium cost.}
Z-Image-Turbo (\$0.005/MP) with CRAFT costs $\approx\$0.044$ per image—over 
$9.5\times$ cheaper than a single Hunyuan Image 3.0 call (\$0.42)—while 
substantially improving quality (DSG1K: VQA $+4.6\%$, DSG $+4.2\%$, Auto SxS 
win rate $+40.0$ p.p.).

\paragraph{Low-cost generators see largest relative gains.}
FLUX-Schnell (\$0.003/MP) achieves the largest quality improvements 
(VQA $+10.3\%$, DSG $+9.0\%$, Auto SxS $+53.4$ p.p.) at minimal absolute cost 
($\approx\$0.029$ total, $2.2\times$ baseline). Here, reasoning overhead 
($\approx 8.3\%$ of total cost) becomes more visible, suggesting adaptive 
early stopping or cheaper verifiers would be particularly beneficial.

\paragraph{Reasoning overhead is negligible for mid-to-high-cost generators.}
For FLUX-Dev (\$0.025/MP), Z-Image-Turbo (\$0.005/MP), and FLUX-2 Pro 
(\$0.0179/MP), the fixed reasoning cost represents only $1.1\%$, $5.5\%$, and 
$1.6\%$ of total CRAFT cost, respectively, making the trade-off highly 
favorable.

\paragraph{Premium models benefit from targeted refinement.}
FLUX-2 Pro (\$0.0179/MP), despite having strong baseline performance, shows 
consistent improvements through CRAFT (VQA $+5.1\%$, DSG $+4.6\%$, Auto SxS 
$+12.0$ p.p.). At $\approx\$0.152$ total cost ($2.0\times$ baseline), it delivers 
state-of-the-art quality while remaining $\approx 64\%$ cheaper than a single 
Hunyuan Image 3.0 call.

\paragraph{Latency and iteration budget.}
In our experiments, a single CRAFT iteration (generation or editing) takes on
average $\approx 30$ seconds, and three iterations are typically sufficient to
reach strong qualitative results. By comparison, a single Hunyuan Image 3.0
generation takes $\approx 60$ seconds on average while delivering lower
compositional quality than CRAFT-equipped mid-cost backbones under our evaluation.

\begin{table}[!h]
    \centering
    \small
    \setlength{\tabcolsep}{5pt}
    \caption{Cost--quality trade-off at $2048{\times}2048$ resolution. CRAFT cost 
    includes two generations plus \$0.0024 reasoning overhead. Cost multiplier shown 
    in parentheses. Quality gains on DSG1K: VQA and DSG as relative improvement (\%); 
    Auto SxS as absolute difference in percentage points (p.p.). Best gains per column 
    in \textbf{bold}.}
    \label{tab:cost_quality_2048}
    \vspace{0.3em}
    \begin{tabular}{@{}lcccccc@{}}
    \toprule
    \multirow{2}{*}{\textbf{Backbone}} & \multirow{2}{*}{$\$/\text{MP}$} &
    \multicolumn{2}{c}{\textbf{Cost @ $2048{\times}2048$}} &
    \multicolumn{3}{c}{\textbf{Quality Gain}} \\
    \cmidrule(lr){3-4}\cmidrule(lr){5-7}
    & & Baseline & CRAFT (2 iters) & $\Delta$VQA (\%) & $\Delta$DSG (\%) & $\Delta$Auto SxS (p.p.) \\
    \midrule
FLUX-Schnell   & 0.0030  & \$0.013 & \$0.029 (2.2$\times$) & \textbf{+10.3\%} & \textbf{+9.0\%} & \textbf{+53.4} \\
FLUX-Dev       & 0.0250  & \$0.105 & \$0.212 (2.0$\times$) & \textbf{+13.0\%} & \textbf{+9.9\%} & +31.0 \\
Z-Image-Turbo  & 0.0050  & \$0.021 & \$0.044 (2.1$\times$) & +4.6\% & +4.2\% & +40.0 \\
FLUX-2 Pro     & 0.0179  & \$0.075 & \$0.152 (2.0$\times$) & +5.1\% & +4.6\% & +12.0 \\
    \midrule
    \multicolumn{7}{@{}l@{}}{\footnotesize \textit{Reference (no CRAFT applied):}} \\
    Hunyuan 3.0 & 0.1000 & \$0.42 & --- & --- & --- & --- \\
    \bottomrule
    \end{tabular}
    \end{table}

\subsection{Qualitative Comparison}

\begin{figure}[h!]
\centering
\setlength{\abovecaptionskip}{-3pt}
\includegraphics[width=0.65\textwidth]{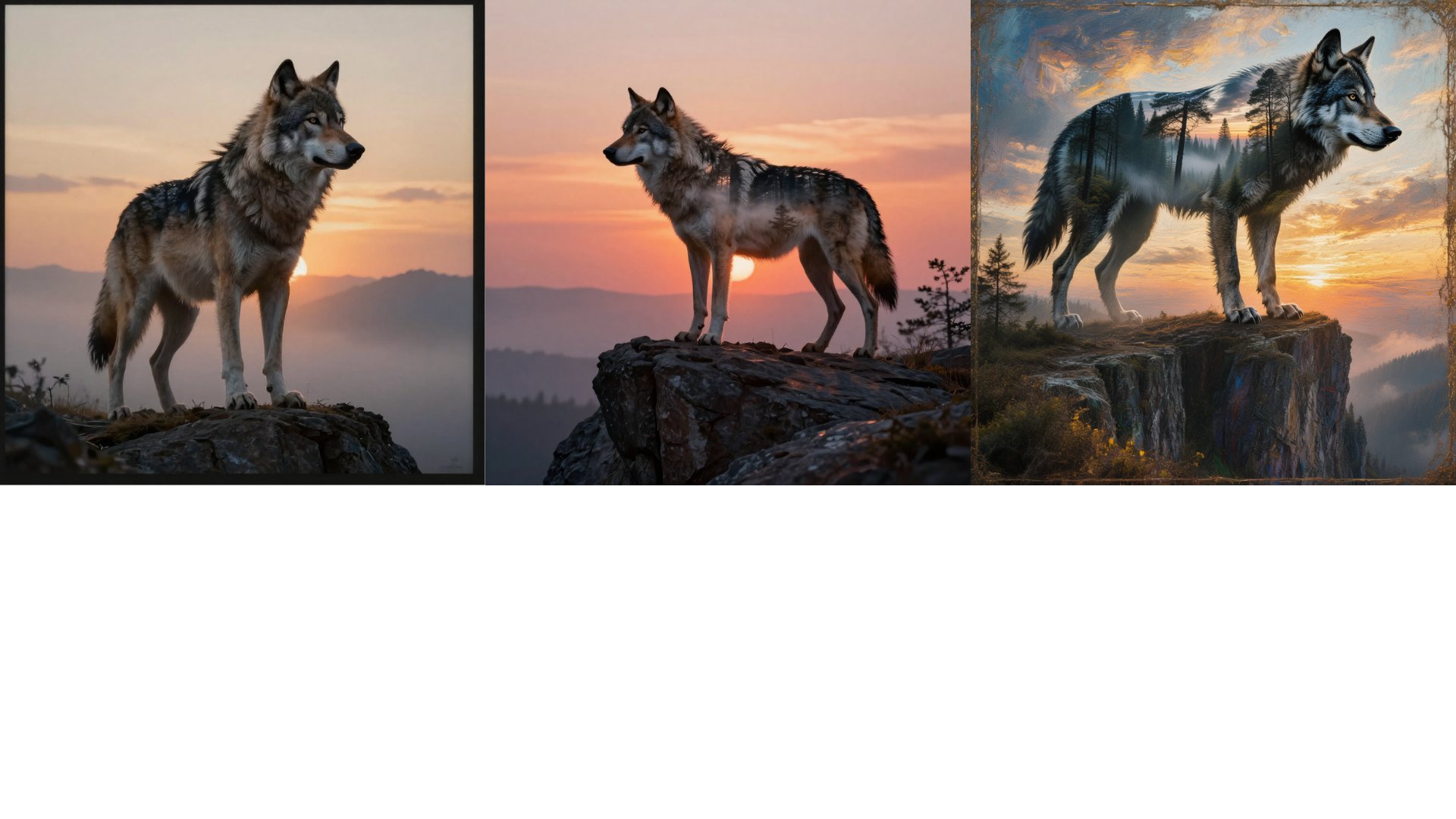}\\[-1em]
\caption{Baseline vs CRAFT vs Hunyuan Image 3 (cost\_prompt1, 
Appendix~\ref{app:cost_prompts}).}
\label{fig:cost1}
\end{figure}

\begin{figure}[h!]
\centering
\setlength{\abovecaptionskip}{-3pt}
\includegraphics[width=0.65\textwidth]{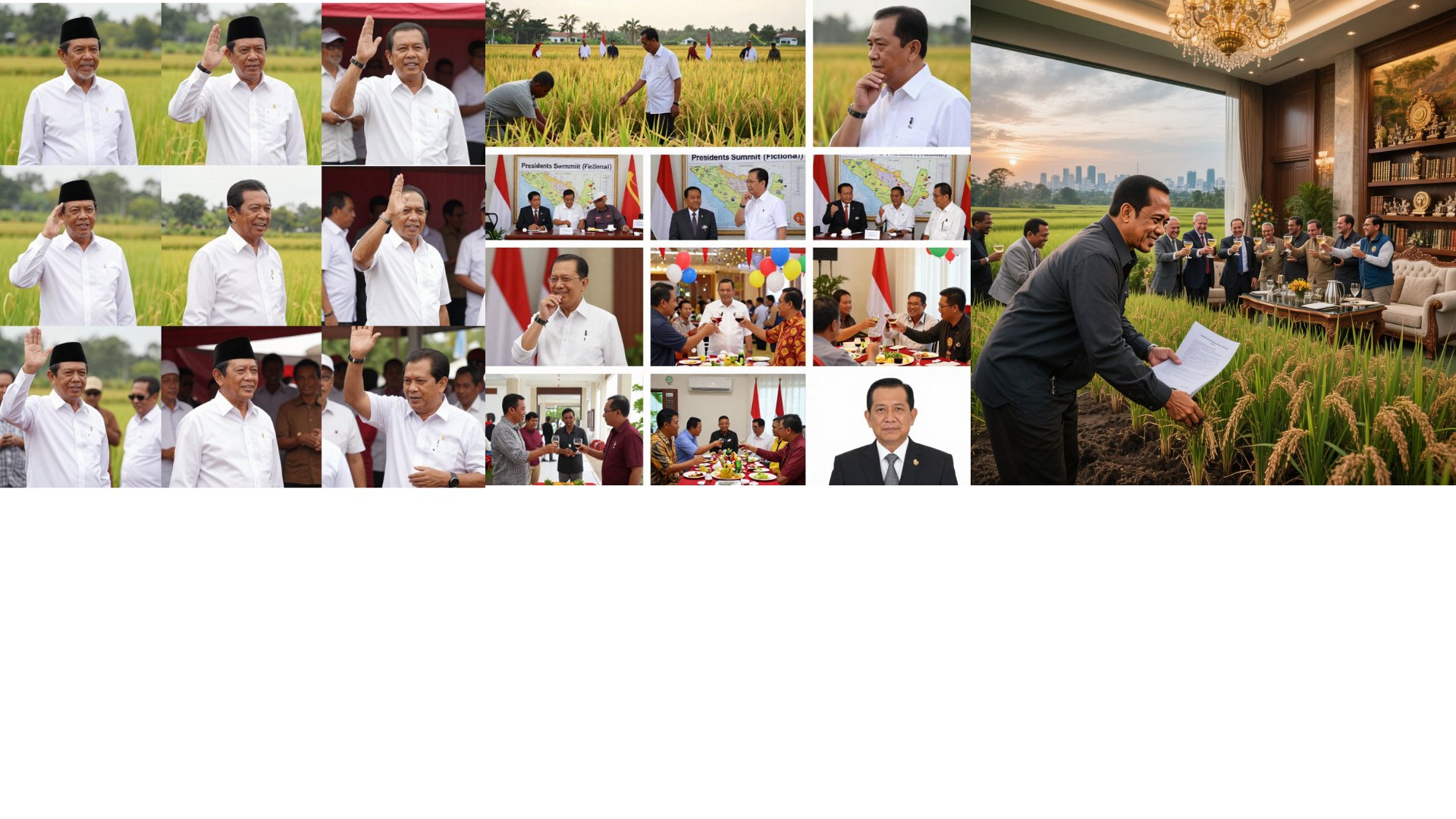}\\[-1em]
\caption{Baseline vs CRAFT vs Hunyuan Image 3 (cost\_prompt2, 
Appendix~\ref{app:cost_prompts}).}
\label{fig:cost2}
\end{figure}

Representative visual comparisons (Figures~\ref{fig:cost1}--\ref{fig:cost2}) 
between Z-Image-Turbo baseline, Z-Image-Turbo with CRAFT, and Hunyuan Image 3.0 
demonstrate that reasoning-guided refinement substantially improves 
object--attribute correctness and spatial relationships, often narrowing the 
gap between lightweight and heavyweight models. Full prompts are provided in 
Appendix~\ref{app:cost_prompts}. Additional examples are available in 
Appendix~\ref{app:additional_cost_comparison}.

\clearpage
%==============================================================================
% Section: Limitations
%==============================================================================

\section{Limitations}
\label{sec:limitations}

Despite its generality, the proposed CRAFT has several limitations. First, the 
refinement loop depends on the reliability of the multimodal LLM responsible for 
visual question evaluation, text recognition, and side-by-side judgment. Although 
stronger MLLMs reduce noise, failures in text reading, misinterpretation of 
fine-grained attributes, or aesthetic biases can propagate incorrect supervisory 
signals across iterations. This is particularly evident for prompts with dense 
textual content, where OCR errors or hallucinated legibility assessments may 
mislead the update step.

Second, while the DVQ framework covers compositional, textual, and 
artifact-related constraints, it still requires that the prompt be mappable to 
a well-defined question set $\mathcal{Q}$. Highly abstract or underspecified 
prompts -- commonly present in datasets such as Parti-Prompt -- remain 
challenging, even after filtering, because they lack concrete visual targets 
for iterative correction.

Third, inference-time cost remains higher than single-shot generation: each 
iteration requires multiple MLLM and T2I calls. Although most gains materialize 
within one or two iterations, the overhead may be non-negligible for large-scale 
or low-latency deployments.

Fourth, evaluation of text rendering quality -- while improved via NED -- 
remains limited by OCR fidelity on stylized or specific-language non-Latin typography.

Finally, while human evaluations (Section~\ref{sec:human_eval}) confirm the 
preference trends observed with automatic side-by-side metrics, the study 
remains small-scale.

%==============================================================================
% Section: Conclusion
%==============================================================================

\section{Conclusion}
\label{sec:conclusion}

We introduced \textbf{CRAFT}, a training-free and model-agnostic reasoning framework
that improves text-to-image and image-editing systems through explicit,
constraint-driven verification and targeted prompt refinement at inference time.
Across all evaluated architectures--from lightweight generators to high-end
diffusion models--the method consistently enhances compositional accuracy,
textual fidelity, artifact detection, and multimodal preference behavior. Notably,
weaker backbones often approach the performance of substantially stronger models
after only one or two reasoning iterations, indicating that structured
inference-time reasoning can partially compensate for architectural limitations
without retraining.

A central advantage of the framework is its full modularity. CRAFT operates as a
plug-in layer that can be paired with any text-to-image generator, any
vision--language model for verification, and any large language model for prompt
rewriting, without model-specific adaptation. As a result, the system directly
benefits from improvements in underlying components, supports heterogeneous
generation pipelines, and can be integrated with minimal engineering effort in
black-box deployment settings.

Because the approach relies on explicit verification, targeted correction, and a
clear stopping criterion, it is not inherently restricted to the image domain.
The same principles suggest promising directions for extending structured,
constraint-driven inference-time reasoning to other multimodal generative settings,
such as video generation and editing (via temporal constraints) or interactive
content generation, provided that intermediate outputs can be evaluated and
corrected in a principled manner.

From a practical standpoint, the reasoning loop introduces negligible cost overhead
relative to modern high-resolution text-to-image models. For example, running two
reasoning iterations for Z-Image-Turbo costs approximately \$$0.0024$ per image,
corresponding to less than $0.3\%$ of the generation cost of Hunyuan Image~3.0 at
comparable resolution. Despite operating at more than $2\times$ lower cost per
megapixel, Z-Image-Turbo equipped with CRAFT often matches or exceeds the
compositional performance of Hunyuan Image~3.0 in our evaluations, demonstrating
that inference-time reasoning offers a cost-efficient alternative to scaling model
size or relying on more expensive diffusion backbones.

Overall, our results show that inference-time reasoning becomes reliable and
scalable when it is made explicit, structured, and verifiable. By grounding
refinement in constraint satisfaction rather than opaque holistic scores, CRAFT
provides a practical and interpretable path toward improving prompt adherence and
reducing failure modes in multimodal generative systems. We hope this work
encourages further exploration of structured reasoning paradigms as a foundation
for more controllable and dependable generative models.

% Bibliography
\bibliographystyle{unsrt}
\bibliography{references}

% Appendix
\clearpage
\appendix
%==============================================================================
% Appendix
%==============================================================================

\section{System Prompts}
\label{app:system_prompts}

This section provides the complete system prompts used for each component of
the CRAFT pipeline. All prompts use GPT-5-nano (gpt-5-nano-2025-08-07) with
temperature=0.0 for reproducibility.

%------------------------------------------------------------------------------
% Initial Prompt Rewriting
%------------------------------------------------------------------------------
\subsection{Initial Prompt Rewriting}
\label{app:prompt_rewrite_init}

\textbf{System Prompt:}

\begin{promptbox}
I want to create an image by giving a prompt to a text-to-image model.

Here is my current prompt: \{ current\_prompt \}

Please generate an improved prompt based on the following guidelines for text-to-image generator:

1. Make sure the prompt contains the elements of subject, context, and style. Specifically:
    - Subject: The main object, person, animal, or scenery you want an image of.
    - Context and background: Also consider the background or environment for the subject. For example, a studio, outdoor scenery, or an indoor setting.
    - Style: Add the style of image you want. Styles can be general (painting, photograph, sketches) or very specific (pastel painting, charcoal drawing, isometric 3D).
        * IMPORTANT: If the original prompt does not explicitly specify a style, add "realistic style" by default in the improved prompt.

2. Use descriptive language: Employ detailed adjectives and adverbs to paint a clear picture for image generator.

3. Provide context: If necessary, include background information to aid the AI's understanding.

4. Reference specific artists or styles: If you have a particular aesthetic in mind, referencing specific artists or art movements can be helpful.

5. Enhancing facial details in images of people:
    - Specify facial details as a focus of the photo (for example, use the word "portrait" in the prompt).

6. CRITICAL - Preserving text in quotation marks:
    - If the original prompt contains text within quotation marks (e.g., "Hello World"), this is LITERAL TEXT that must appear in the generated image.
    - You MUST preserve this text EXACTLY as written, including the quotation marks.
    - DO NOT modify, rephrase, translate, or remove any text that appears in quotation marks.
    - DO NOT remove the quotation marks themselves - they indicate visible text.
    - Example: 'sign that says "Welcome"' -> keep "Welcome" exactly as is with quotes
    - Example: 'banner with text "Sale 50%"' -> keep "Sale 50%" exactly as is with quotes

7. IMPORTANT - Handling titles vs visible text:
    - If the prompt contains "with the title X", "titled X", "named X", or "called X" WITHOUT quotation marks - this refers to the ARTWORK'S NAME, NOT text that should appear in the image.
    - Rewrite these as style/theme descriptions without implying visible text. 
    - Example: "with the title Karakuri Ninja Oboro" -> "depicting a scene from Karakuri Ninja Oboro" or "featuring the Karakuri Ninja Oboro theme"
    - Only text explicitly in quotation marks should be treated as visible text in the image.

8. Avoid ambiguity about text:
    - Text WITH quotation marks = visible text in the image (preserve exactly)
    - Text WITHOUT quotation marks = description/style/theme (can be rephrased)
    - Never add quotation marks unless they were in the original prompt
    - Never remove quotation marks if they were in the original prompt

Please now generate 1 improved prompt and enclose it between <PROMPT> and </PROMPT>. 
If you believe the current prompt is already satisfactory, simply respond with "<PROMPT> NO\_CHANGE </PROMPT>".

\end{promptbox}

%------------------------------------------------------------------------------
% Verify and Self Correct
%------------------------------------------------------------------------------
\subsection{Verify and Self Correct}
\label{app:prompt_verify_and_self_correct}

\textbf{System Prompt:}

\begin{promptbox}
    Your task is to first verify whether my original prompt satisfies each of the constraints specified by the user: \{ constraints \}
Make sure you verify each constraint and answer "YES" if the constraint is met or "NO" otherwise, followed by a short explanation. 
If there is any unmet constraint, please revise the prompt to satisfy all constraints. 
Make sure you only modify the part of the prompt that previously led to constraint violations and bracket any revised
prompt between <answer> and </answer>. If the original prompt satisfies all constraints, you may simply write
"<answer> NO_CHANGE </answer>".
\end{promptbox}

%------------------------------------------------------------------------------
% VQA Evaluation
%------------------------------------------------------------------------------
\subsection{VQA Evaluation}
\label{app:prompt_vqa}

\textbf{System Prompt:}

\begin{promptbox}
Answer only with "Yes" or "No". Do not give other outputs or punctuation marks.
    Image: \{ image \}
    Question: \{ dvq[i] \}
\end{promptbox}

%------------------------------------------------------------------------------
% DVQ Rationalization
%------------------------------------------------------------------------------
\subsection{DVQ Rationalization}
\label{app:prompt_dvq_rationalization}

\textbf{System Prompt:}

\begin{promptbox}
    You are an expert in examining images generated by text-to-image generation models. The user specifies a list of questions, each
of which contains a property or characteristic that the generated image should have and the goal is to make sure that when an
reviewer reviews the generated image, they will answer "Yes" to each of these questions. However, the reviewer answered "No" to
the question below to the current image:
Image: \{ image \}
Question: \{ dvq[i] \}
\end{promptbox}

%------------------------------------------------------------------------------
% Targeted Prompt Refinement
%------------------------------------------------------------------------------
\subsection{Targeted Prompt Refinement}
\label{app:prompt_refinement}

\textbf{System Prompt:}

{\footnotesize\ttfamily\raggedright\sloppy
\begin{quote}
You are an expert in prompt engineering for text-to-image generation models, which take a text prompt as input and generate images depicting the prompt as output. Your task is to improve upon my current prompt based on feedback I obtained from human reviewers.

The user has specified a list of questions, each of which contains a property or characteristic that the generated image should have and the goal is to make sure that when a reviewer reviews the generated image, they will answer "Yes" to all the questions. However, they have answered "No" to one or more questions and provided some feedback, and I'd like you to help me improve my existing prompt, taking into account their feedback.

My current prompt is: '\{best\_prompt\_so\_far\}'

But a reviewer looked at my image and answered "No" to the following question(s). Below are the questions and their feedback, which contains explanations why they believed the image does not satisfy the requirement in the question(s) and their suggestions for improvements:

\{feedback\_section\}

Based on your analysis and the reviewer's opinion, now improve my prompt by incorporating all their feedback points into new prompt(s) so that the reviewer will now answer "YES" to the question(s) when I generate a new image with the new prompt(s). Pay close attention to any points that they mentioned are lacking in the current image, and ensure you incorporate all these information in your revised prompts. Make sure the revised prompts are concise and effective. You may be creative in composing the new prompts, adding new elements, emphasize existing elements, or shuffle the order of the sentences to emphasize the previously missing characteristics at the beginning of the prompt. Finally, make sure each prompt you generated is bracketed between <PROMPT> and </PROMPT>
\end{quote}
}

%------------------------------------------------------------------------------
% Side-by-Side Judge
%------------------------------------------------------------------------------
\subsection{Side-by-Side Judge}
\label{app:prompt_judge}

\textbf{System Prompt:}

\begin{promptbox}
    You are an expert in critiquing images. Given a user prompt, a text-to-image model generated two images
("Image A" and "Image B"), and your task is to pick the overall better image. 
The user prompt is: "\{user\_prompt\}"
Image A: \{image\_A\}
Image B: \{image\_B\}
Please carefully explain the rationale before giving your final choice. Make sure you finish your answer with
"<answer> X </answer>", where X is "A" or "B".
\end{promptbox}

%------------------------------------------------------------------------------
% Image Editing
%------------------------------------------------------------------------------
\subsection{Image Editing}
\label{app:prompt_editing}

\textbf{System Prompt:}

{\footnotesize\ttfamily\raggedright\sloppy
\begin{quote}
You are an expert in prompt engineering for image editing models, which take an instruction prompt and an existing image as input and produce an edited image as output. Your task is to improve my current editing instruction based on feedback I obtained from human reviewers.

The user has specified a list of questions, each describing a property or characteristic that the edited image should have. The goal is to ensure that, when a reviewer examines the edited image, they will answer 'Yes' to all questions regarding these requirements. However, the reviewer answered "No" to one or more questions and provided feedback explaining why the image did not satisfy these requirements, along with suggestions for improvement.

My current image editing prompt is: '\{best\_prompt\_so\_far\}'

Below are the questions and the reviewer's feedback, outlining what is lacking in the current edited image and any suggested improvements:

\{feedback\_section\}

Based on your analysis and the reviewer's feedback, improve my prompt by incorporating all their feedback and suggestions. Your revised editing instruction(s) should address every missing characteristic so that the reviewer will now answer 'Yes' to every question when a new edit is made using your new prompt. Carefully consider all points mentioned as lacking and make sure they are explicitly incorporated in your improved prompts. Feel free to creatively rephrase, reorder, or emphasize specific elements to ensure clarity and effectiveness. Each improved instruction should be concise, clear, and actionable. Finally, enclose each improved prompt between <PROMPT> and </PROMPT>.
\end{quote}
}

\clearpage

%==============================================================================
% Long-Form Prompts
%==============================================================================

\section{Long-Form Prompts}
\label{app:long_prompts}

\subsection{long\_prompt1}

``Studio shot of modern cosmetic bottles on textured marble and velvet
background, soft golden lighting, luxury minimalist style, 8K, hyperrealistic.
Text elements: large header `LUMINOUS SKIN' in a thin geometric sans-serif font
(like Futura PT Light) in the top left corner, a central mathematical fraction
`V/C = $\infty$' in an elegant serif font (like Garamond) slightly off-center
to the right, the caption `where Value meets Clarity' in a thin italic serif
font beneath the fraction, and a tagline `The future of radiance is here.' in a
light sans-serif font (like Montserrat Light) in the bottom right corner. Color
palette: cream, charcoal, dark khaki, gold accents.''

\subsection{long\_prompt2}

``Studio shot of a classic wooden chess set displayed in multiple angles on a
clean white background, soft diffused daylight, product photography style, 8K,
ultra-realistic. The main composition features a natural wood chessboard with
alternating maple-light and walnut-dark squares, bordered by engraved
coordinates in white. In the foreground, the board is fully set up for a game
with traditional Staunton-style pieces in cream and ebony tones, arranged in
perfect symmetry. Supporting shots above show the same board folded open (flat
top view) and closed (vertical orientation), emphasizing craftsmanship and
portability. Lighting is neutral and shadow-soft, highlighting the wood grain
texture and subtle gloss finish. Text elements: Large minimalist header
`CLASSIC CHESS SET' in uppercase geometric sans-serif (e.g. Futura PT Medium)
centered at the top; Sub-caption `Precision. Tradition. Balance.' in light
italic serif (e.g. Garamond Italic) beneath the header; Tagline `Handcrafted
strategy for every mind.' in thin sans-serif (e.g. Montserrat Light) in the
lower right corner. Color palette: natural walnut brown, ivory, maple beige,
white background, soft gray shadows. Mood: timeless elegance, craftsmanship,
calm intellectual focus.''

\subsection{long\_prompt3}

``Studio shot of a luxury single-watch travel case and timepiece on a seamless
white background, minimalist product photography, 8K, ultra-realistic. The
composition features a tan leather cylindrical case with soft suede interior,
opened to reveal its plush texture and metal snap closures. Beside it, a
stainless-steel wristwatch with a silver dial and black numerals rests on a
matching suede-brown cushion, aligned precisely for balance and symmetry.
Lighting: soft neutral daylight with gentle reflections on the watch bezel and
bracelet, creating a clean, premium aesthetic. Shadows are soft and diffuse,
emphasizing material contrast between metal, leather, and suede. Text elements:
Large elegant header `TRAVEL IN TIME' in all caps, geometric sans-serif (e.g.
Futura PT Medium) at the top center; Sub-caption `A refined home for your
watch.' in light italic serif (e.g. Garamond Italic) below; Tagline `Crafted
for journeys. Designed for care.' in thin sans-serif (e.g. Montserrat Light) at
the lower right corner. Color palette: warm caramel brown, beige suede, polished
silver, white background, soft gray shadows. Mood: refined craftsmanship, modern
classic elegance, understated luxury.''

\subsection{long\_prompt4}

``Interior lifestyle shot of a modern bamboo laundry hamper in a minimalist home
setting, 8K hyperrealistic, warm natural lighting. The hamper stands on a light
wooden floor beside a cozy cream armchair draped with a knitted throw and a soft
off-white rug. Its tall, rectangular form features smooth rounded edges,
vertical bamboo slats, and a clean white fabric rim. A twisted rope handle
adorns each side, matching the handle on the lid -- blending natural materials
with functional design. Lighting: soft daylight filtered through sheer curtains,
creating gentle shadows and a serene, homely atmosphere. Composition: centered
object focus, surrounded by neutral-toned decor -- a wooden desk with a
minimalist ceramic vase holding dried flowers, and a framed mirror subtly
visible in the background. Color palette: warm honey bamboo, ivory white, beige,
cream, and soft taupe accents. Mood: calm, clean, eco-conscious aesthetic --
evoking Scandinavian simplicity and natural harmony. Text elements (optional for
layout): Header: `Sustainable Living Starts at Home' in soft sans-serif (e.g.,
Helvetica Neue Light); Caption: `Elegant bamboo design for everyday
organization.' in italic serif (e.g., Garamond Italic); Tagline: `Minimal form.
Maximum calm.' in thin uppercase (e.g., Futura PT Light).''

\subsection{Additional Long-Form Results}

\begin{figure}[!h]
    \centering
    \includegraphics[width=0.8\textwidth]{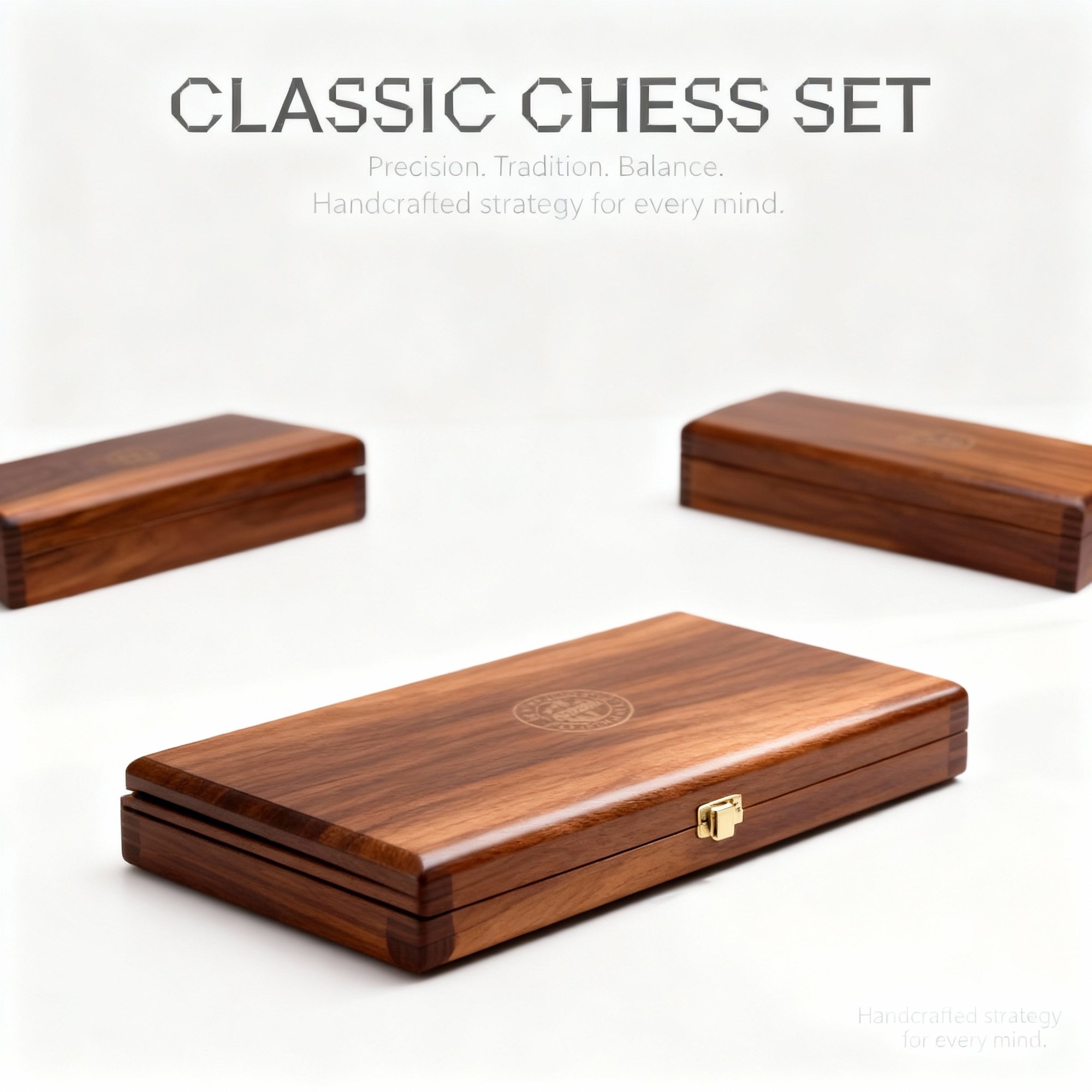}
    \caption{CRAFT results for long\_prompt2.}
    \label{fig:long2_app}
\end{figure}

\begin{figure}[!h]
    \centering
    \includegraphics[width=0.8\textwidth]{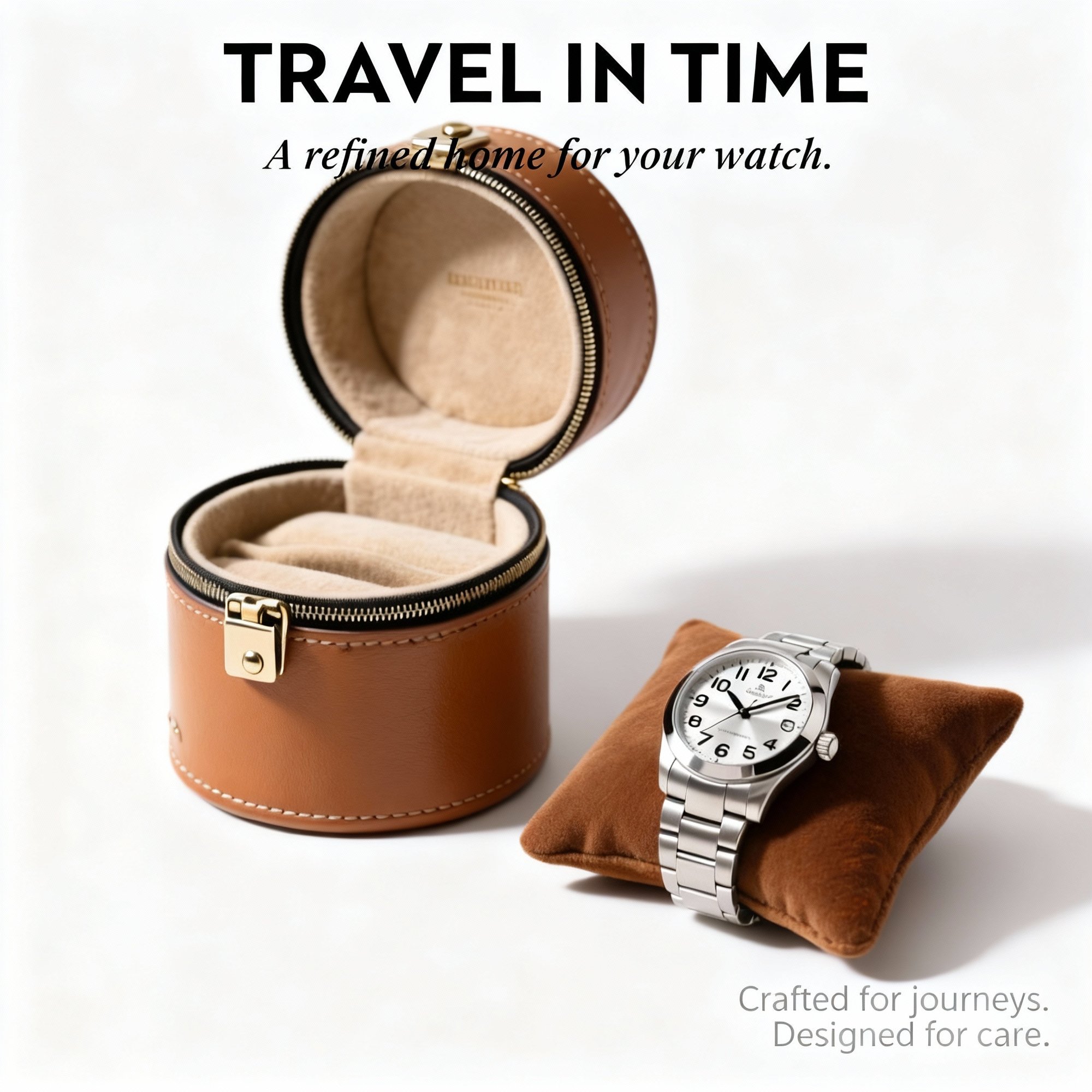}
    \caption{CRAFT results for long\_prompt3.}
    \label{fig:long3_app}
\end{figure}

\begin{figure}[!h]
    \centering
    \includegraphics[width=0.8\textwidth]{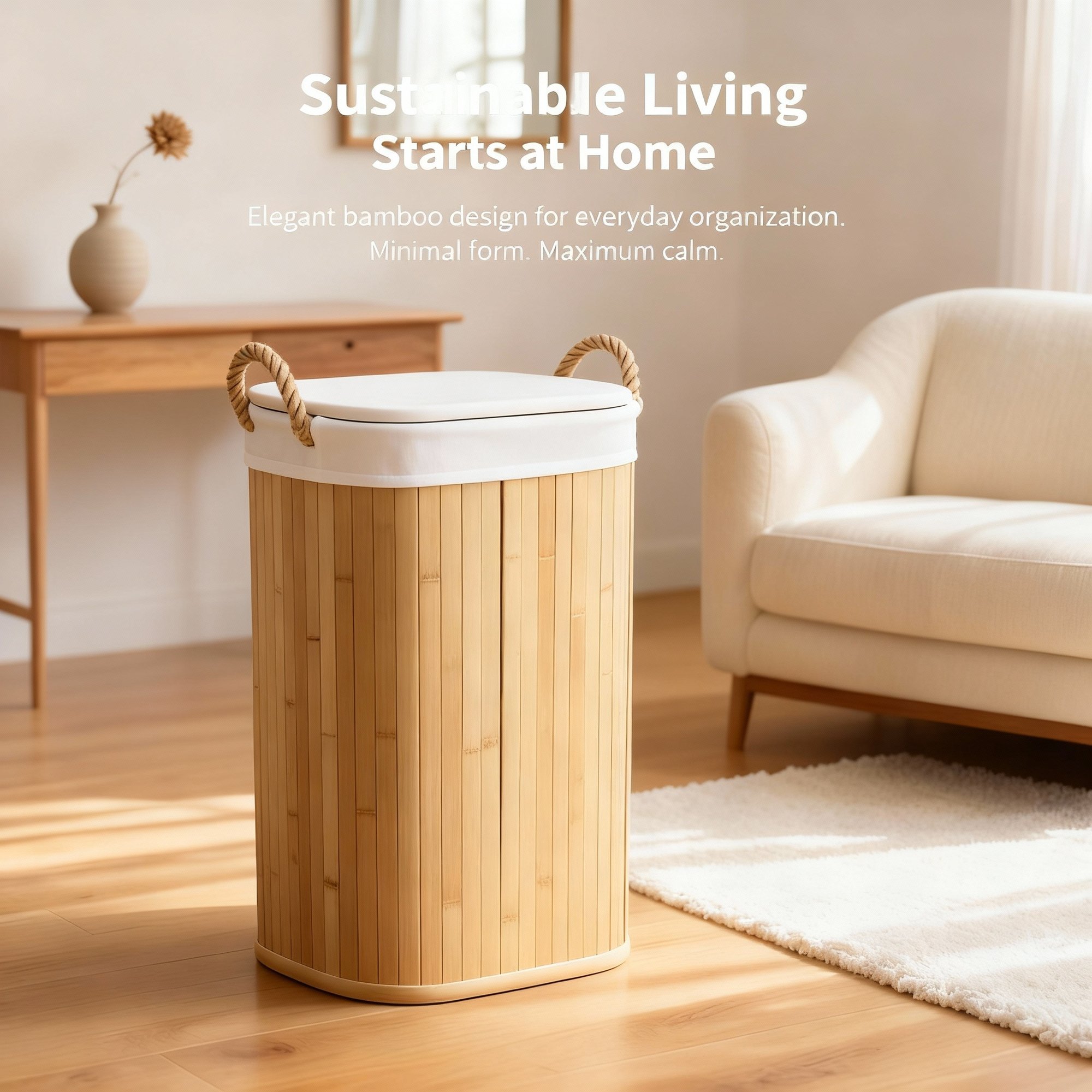}
    \caption{CRAFT results for long\_prompt4.}
    \label{fig:long4_app}
\end{figure}

\clearpage
%==============================================================================
% Cost Comparison Prompts
%==============================================================================

\section{Cost Comparison Prompts}
\label{app:cost_prompts}

\subsection{cost\_prompt1}

``Portrait Pic, mysterious silhouette of a majestic wolf standing on a cliff at
sunset, created with a dreamy double exposure effect showing a misty forest
inside its silhouette, combined with expressive oil brushstrokes for texture and
depth, cinematic lighting, surreal atmosphere, amazing depth, double exposure,
surreal, intricately detailed, perfect balanced, deep fine borders, artistic
photorealism, highly detailed, masterpiece, award winning, double exposure.''

\subsection{cost\_prompt2}

``Generate a series of six candid, documentary-style photos of this Indonesian
president in office, in the rice fields, and partying with other presidents.''

\subsection{cost\_prompt3}

``Leela replaces the original heroine: tall, athletic cyclops woman, purple
ponytail, white tank top, black trousers, shoulder harness, sweat-stained and
battle-worn. She's carrying Nibbler in her left arm: small black three-eyed
alien in red cape and white shorts, clinging to her with worried expression.
Leela grips a detailed sci-fi pulse rifle in one hand, matching the exact shape
and silhouette of the Aliens weapon, but drawn in Futurama's clean cartoon
linework and limited shading. Blue backlight haze, drifting smoke, cinematic
composition, dramatic rim-lighting on Leela's face. Keep Leela and Nibbler true
to Futurama's animation style, smooth outlines, bold colours, simple
cel-shading. Background stays faithful to the original Aliens reference:
industrial sci-fi corridor, diffused fog, blue-grey tones, subtle reflections.
Use both reference images for design accuracy: the pose, framing, weapon and
atmosphere from the Aliens still, and Leela + Nibbler's exact character designs
from Futurama.''

\subsection{cost\_prompt4}

``Create a single cinematic illustration that visually represents the following
poem, capturing its emotions, metaphors, and atmosphere Good night mother, Good
night father, Kiss your little son. Good night sister, Good night brother, Good
night everyone.''

\subsection{cost\_prompt5}

``A hyperrealistic photo of glass-made samgyeopsal placed on a grill.''

\subsection{cost\_prompt6}

``Ultra-realistic Roronoa Zoro portrait.''

\clearpage

%==============================================================================
% Additional Qualitative Results
%==============================================================================

\section{Additional Qualitative Results}
\label{app:additional_qualitative}

This section provides additional qualitative examples for models evaluated in
Section~\ref{sec:qualitative}, demonstrating further improvements achieved by
CRAFT across diverse prompts and backbones.

\subsection{Qwen Image Additional Examples}

\begin{figure}[!h]
    \centering
    \includegraphics[width=0.45\textwidth]{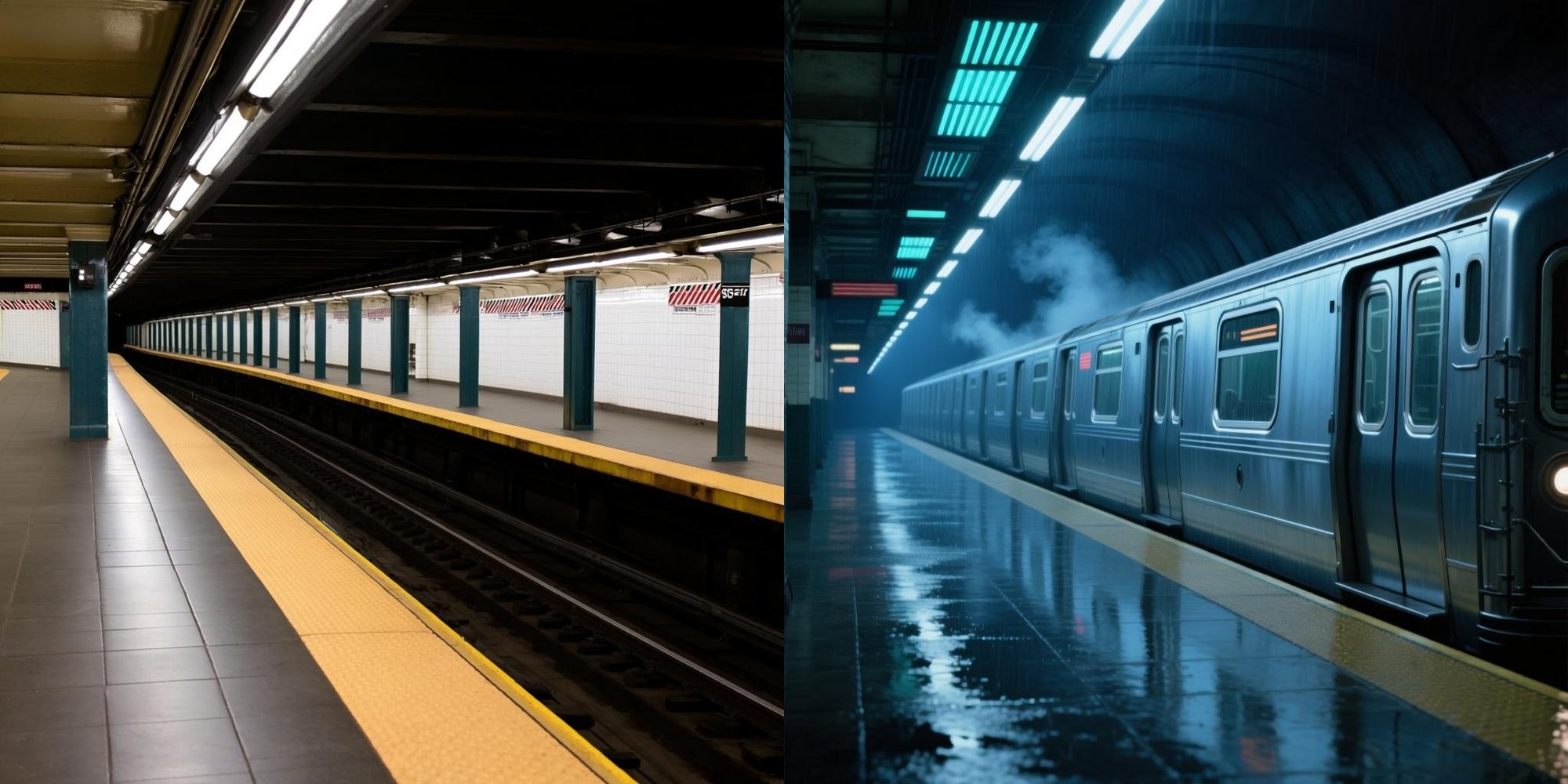}
    \caption{Baseline vs CRAFT. Prompt: \textit{a subway train in an empty
            station}.}
    \label{fig:qual2_app}
\end{figure}

\begin{figure}[!h]
    \centering
    \includegraphics[width=0.45\textwidth]{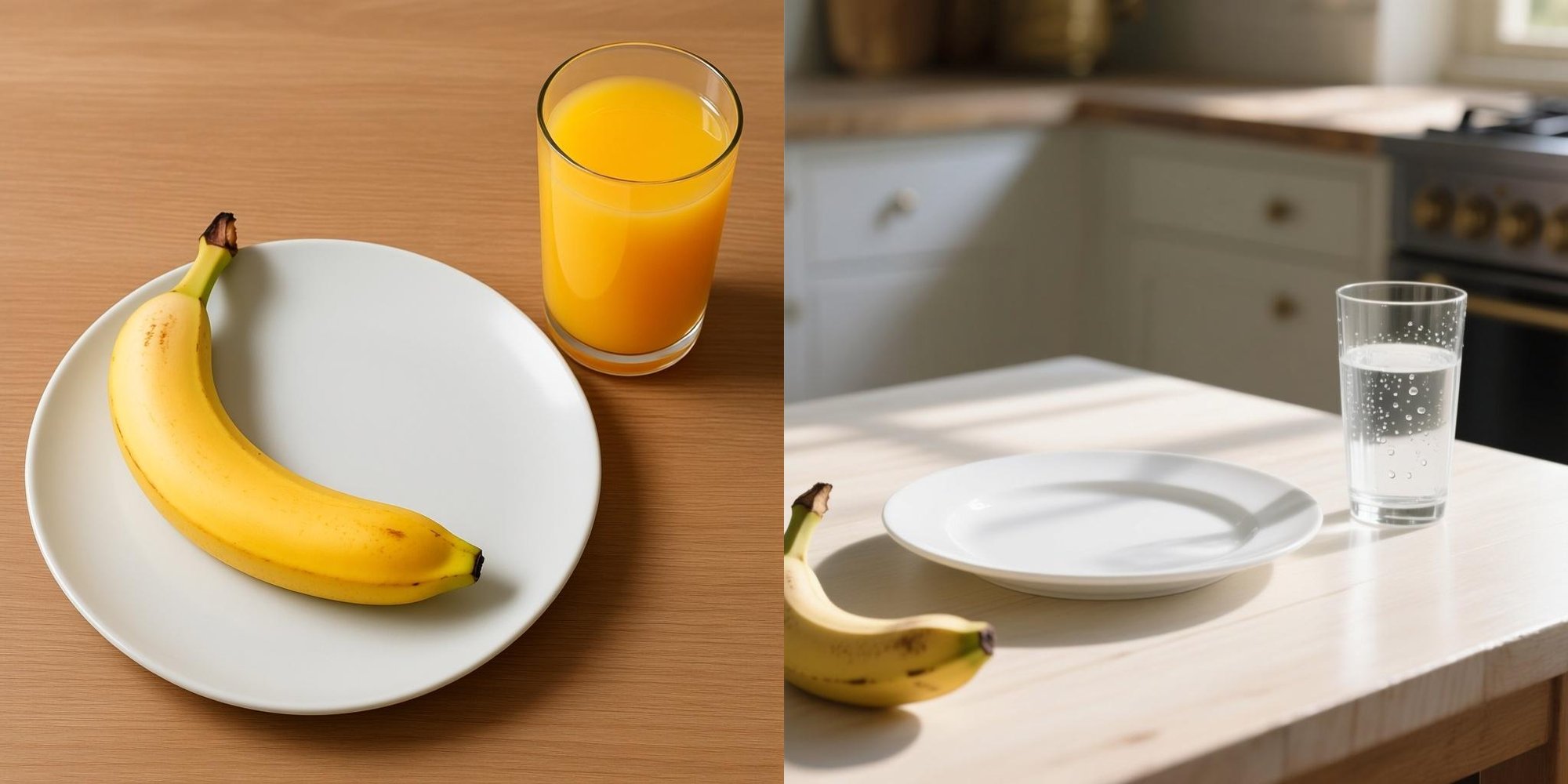}
    \caption{Baseline vs CRAFT. Prompt: \textit{a plate that has no bananas on it.
            There is a glass without orange juice next to it}.}
    \label{fig:qual4_app}
\end{figure}

\begin{figure}[!h]
    \centering
    \includegraphics[width=0.45\textwidth]{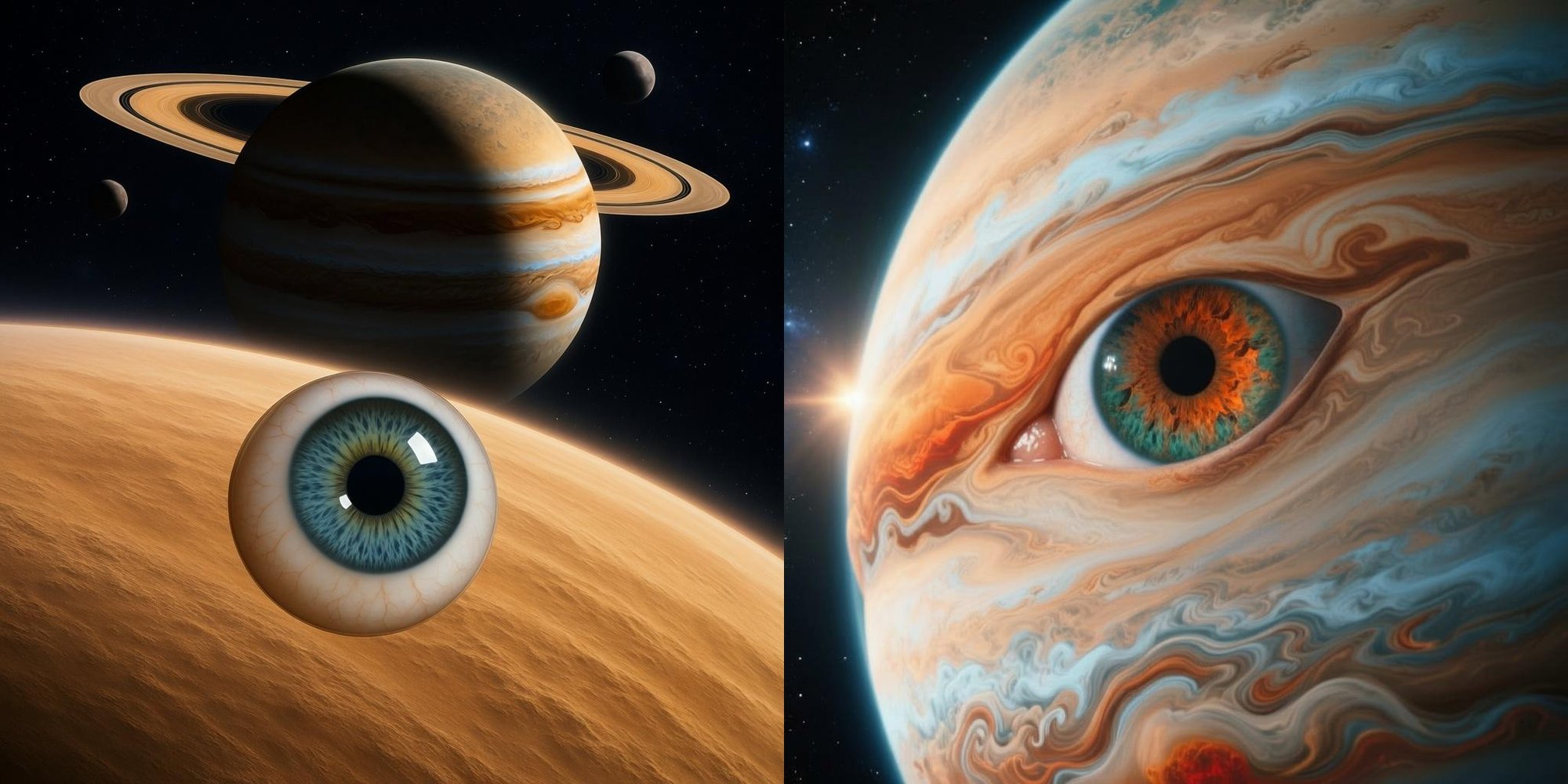}
    \caption{Baseline vs CRAFT. Prompt: \textit{the eye of the planet Jupiter}.}
    \label{fig:qual5_app}
\end{figure}

\clearpage
\subsection{FLUX-Dev Additional Examples}

\begin{figure}[!h]
    \centering
    \includegraphics[width=0.45\textwidth]{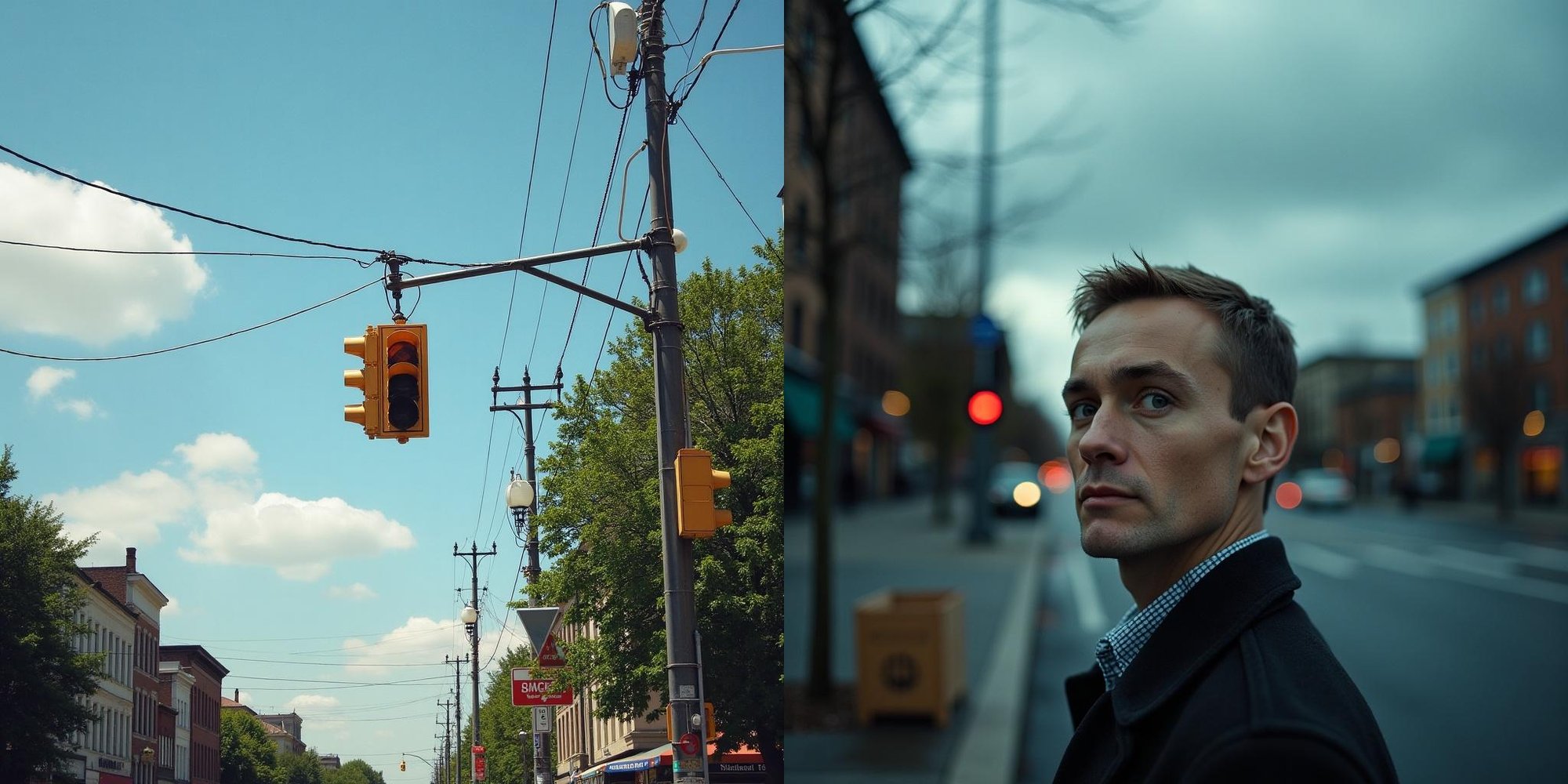}
    \caption{Baseline vs CRAFT. Prompt: \textit{sky above person. sky above street.
            sky above tree. sky above traffic light. sky above box. box on street}.}
    \label{fig:fluxdev4_app}
\end{figure}

\begin{figure}[!h]
    \centering
    \includegraphics[width=0.45\textwidth]{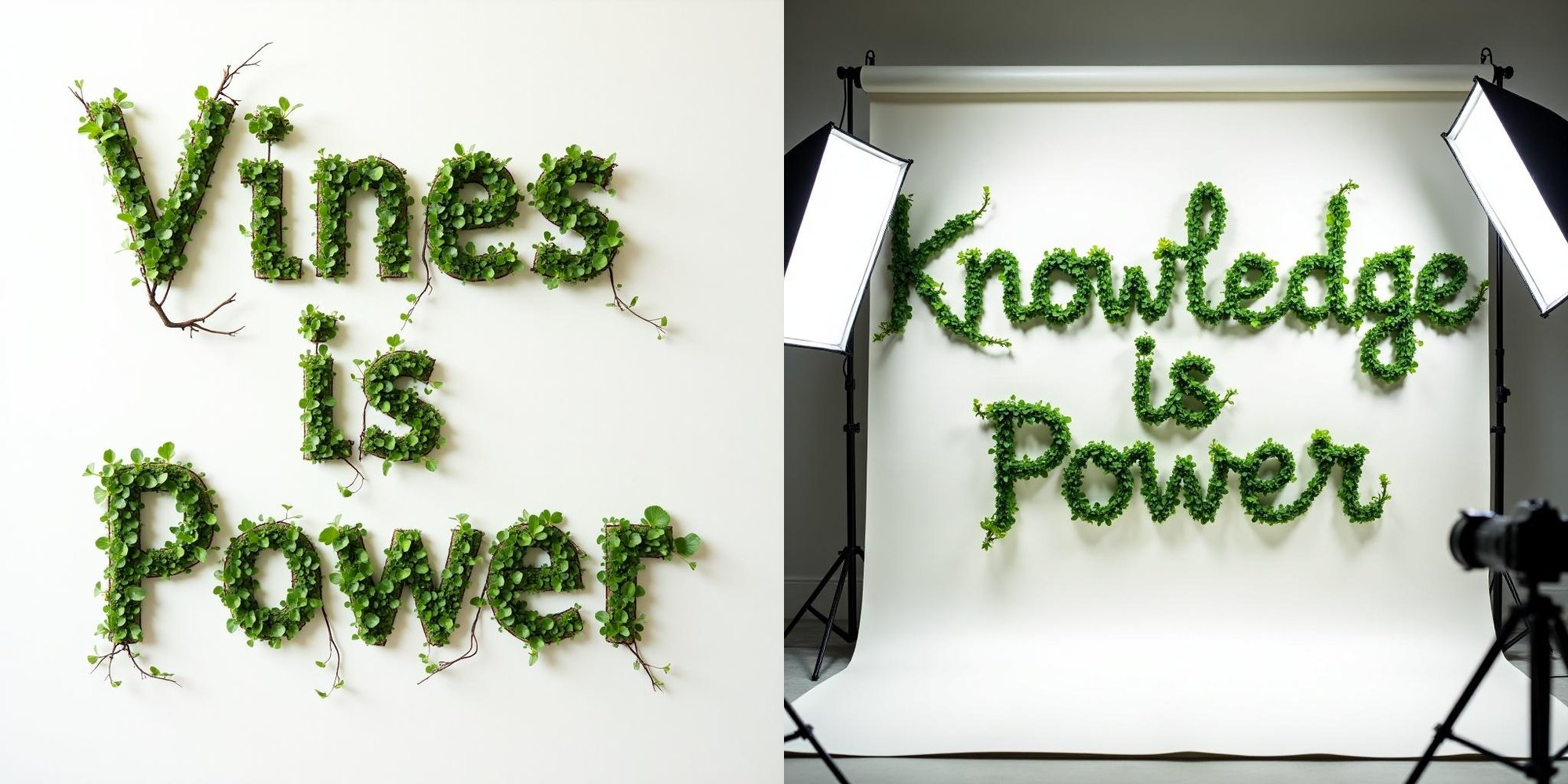}
    \caption{Baseline vs CRAFT. Prompt: \textit{studio shot of vines in the shape
            of text knowledge is power sprouting, centered}.}
    \label{fig:fluxdev5_app}
\end{figure}

\clearpage

\subsection{Z-Image Additional Examples}

\begin{figure}[!h]
    \centering
    \includegraphics[width=0.45\textwidth]{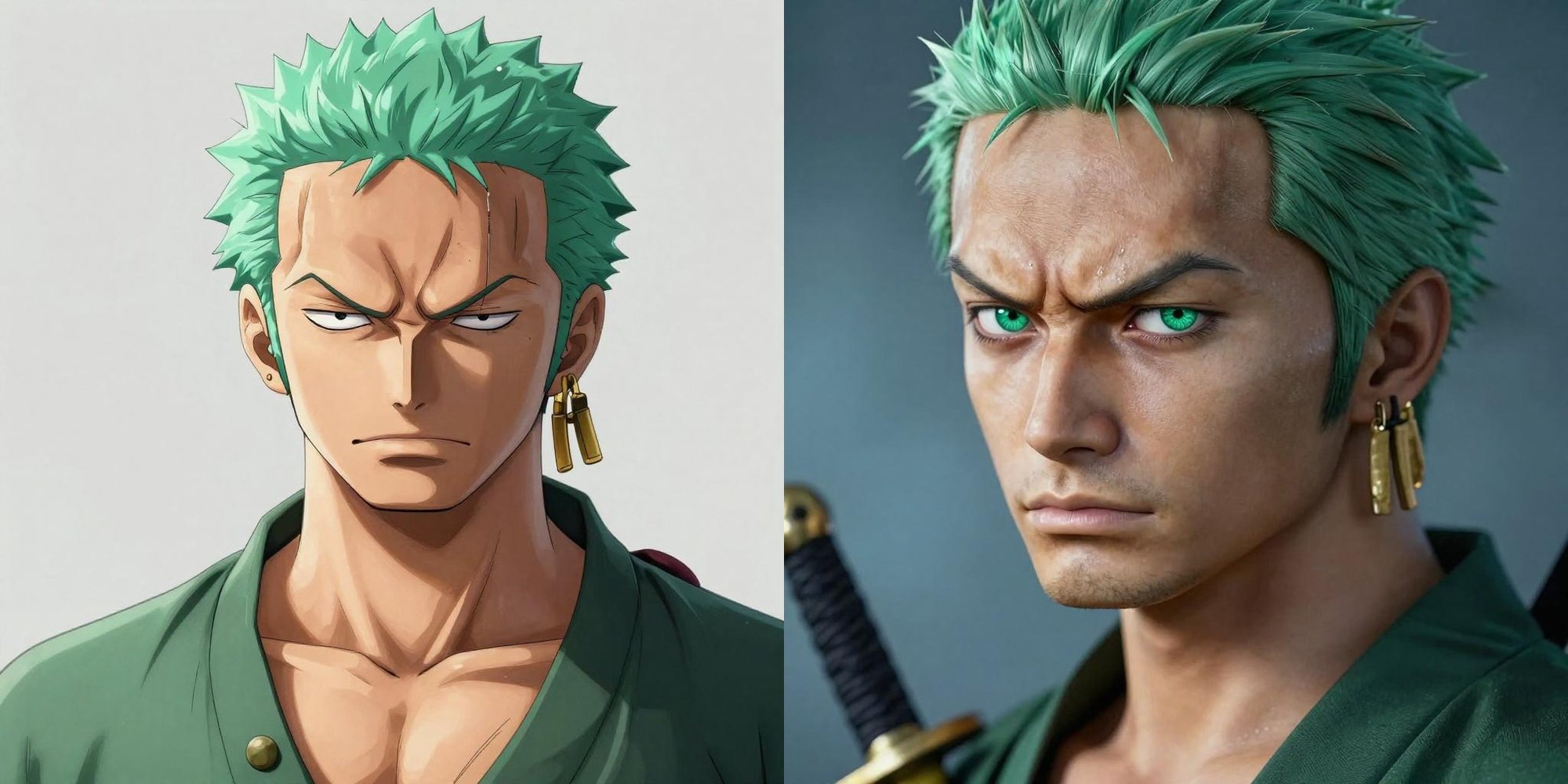}
    \caption{Baseline vs CRAFT. Prompt: \textit{Ultra-realistic Roronoa Zoro
            portrait}.}
    \label{fig:zimage3_app}
\end{figure}

\begin{figure}[!h]
    \centering
    \includegraphics[width=0.45\textwidth]{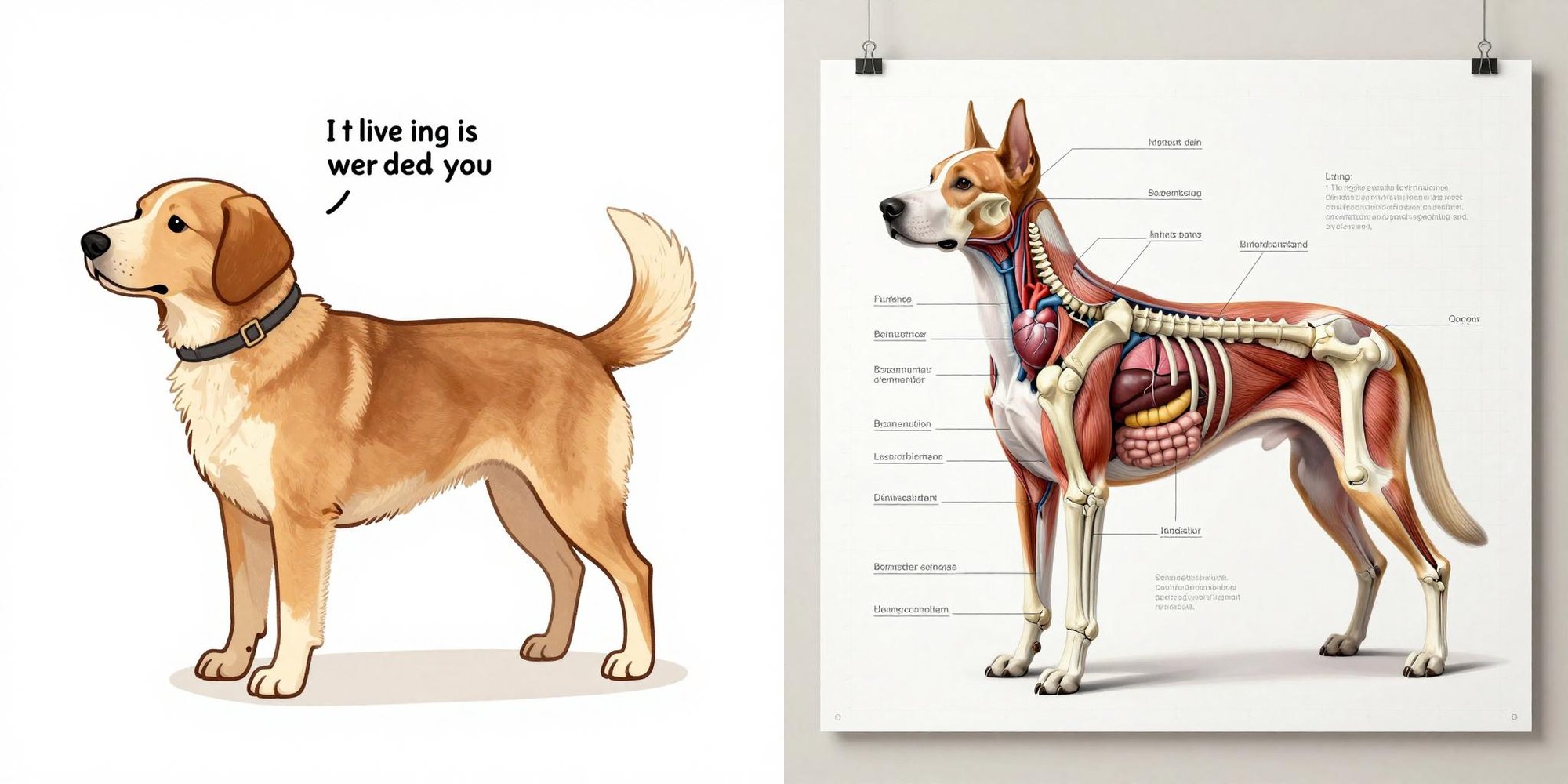}
    \caption{Baseline vs CRAFT. Prompt: \textit{funny anatomical diagram of dog pet,
            with humorous annotations}.}
    \label{fig:zimage4_app}
\end{figure}

\clearpage

\subsection{FLUX-2 Pro Additional Examples}

\begin{figure}[!h]
    \centering
    \includegraphics[width=0.45\textwidth]{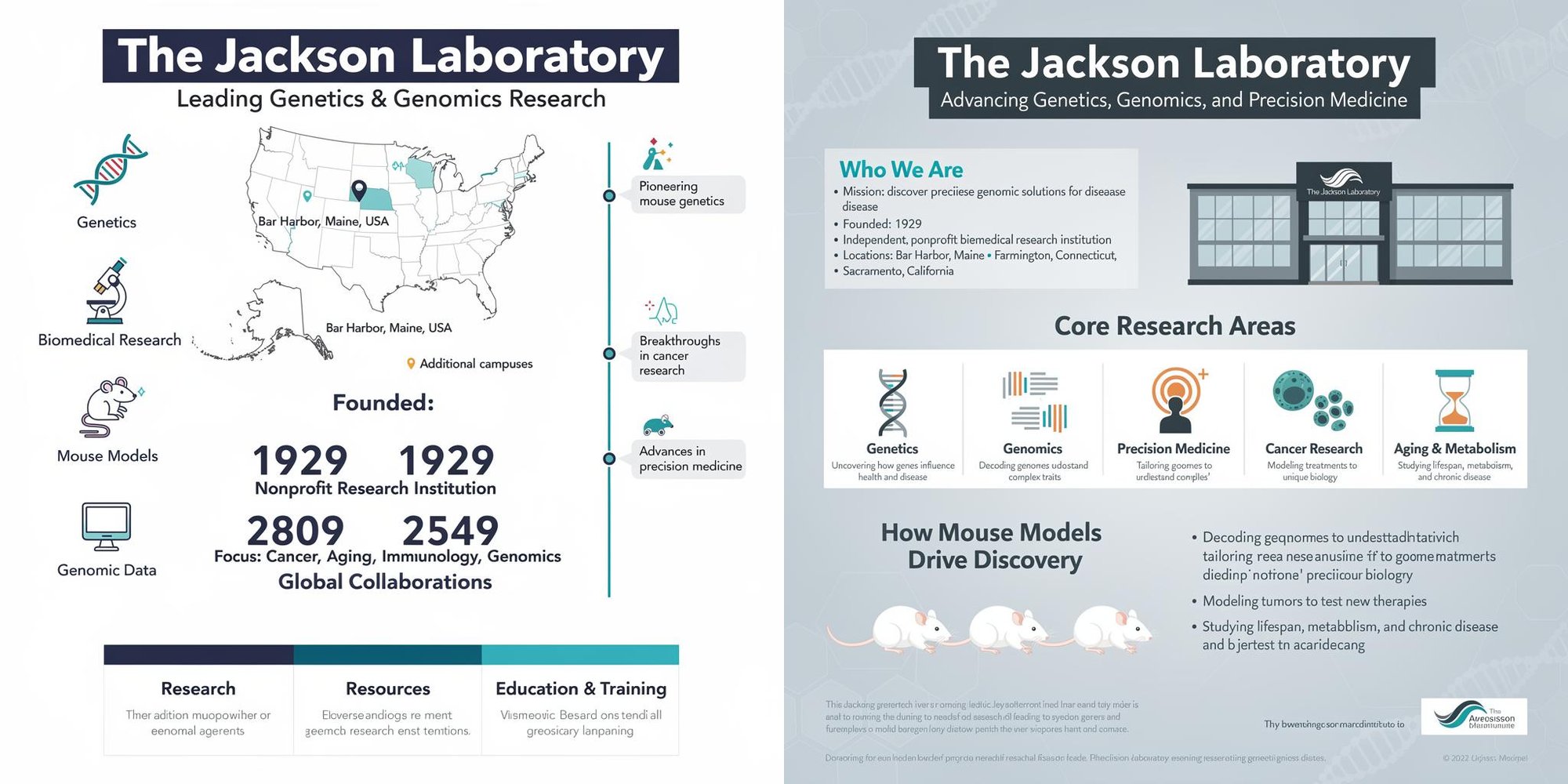}
    \caption{Baseline vs CRAFT. Prompt: \textit{Infographic about the Jackson
            Laboratory en}.}
    \label{fig:flux2pro3_app}
\end{figure}

\begin{figure}[!h]
    \centering
    \includegraphics[width=0.45\textwidth]{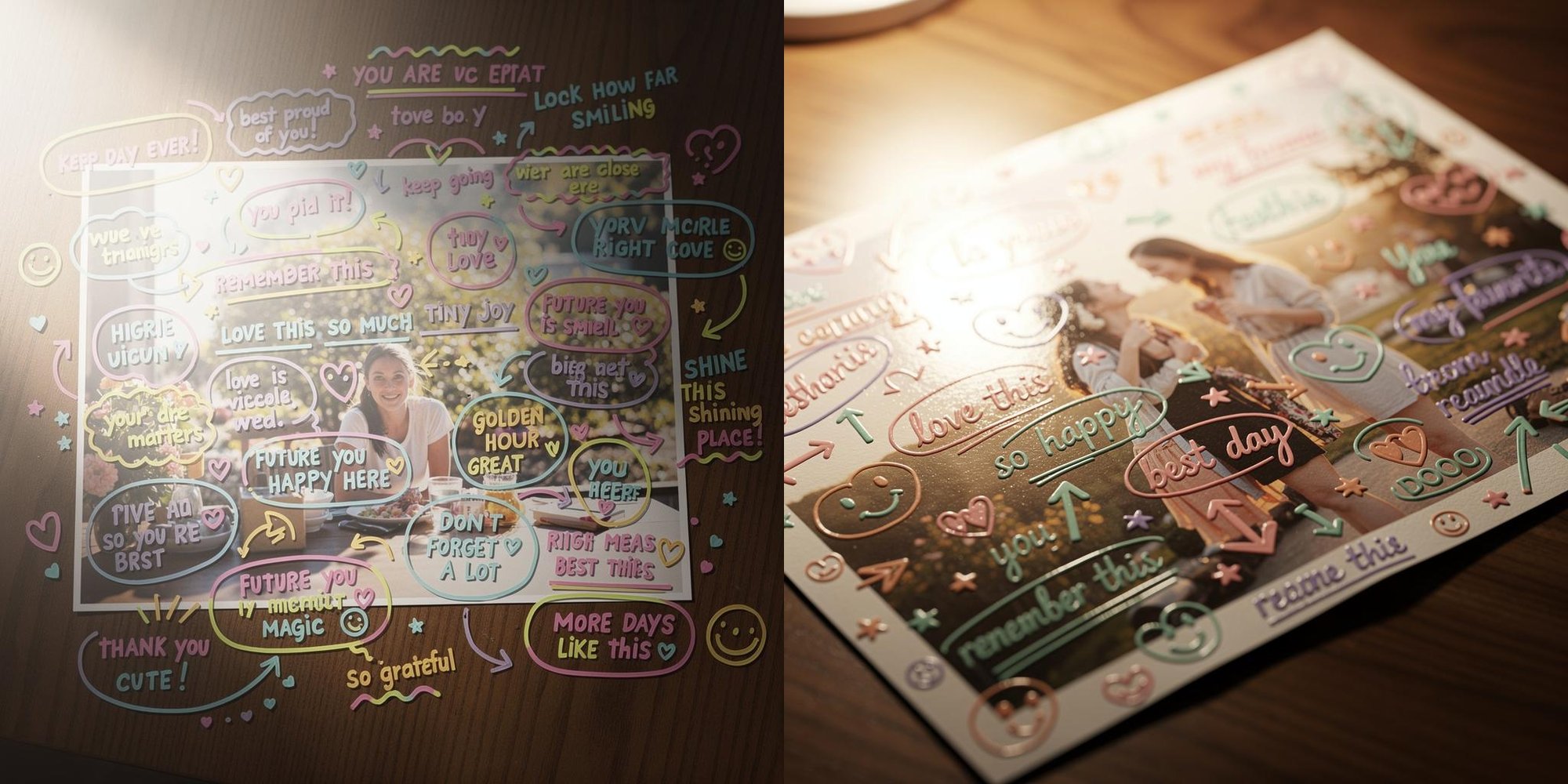}
    \caption{Baseline vs CRAFT. Prompt: \textit{Write as many cute handwritten
            notes and highlights as possible on the photo!}}
    \label{fig:flux2pro4_app}
\end{figure}

\clearpage

%==============================================================================
% Additional Cost Comparison Examples
%==============================================================================

\section{Additional Cost Comparison Examples}
\label{app:additional_cost_comparison}

This section provides additional cost--quality comparison examples between
Z-Image-Turbo baseline, Z-Image-Turbo with CRAFT, and Hunyuan Image 3.0. Full
prompts are available in Section~\ref{app:cost_prompts}.

\begin{figure}[!h]
    \centering
    \includegraphics[width=0.65\textwidth]{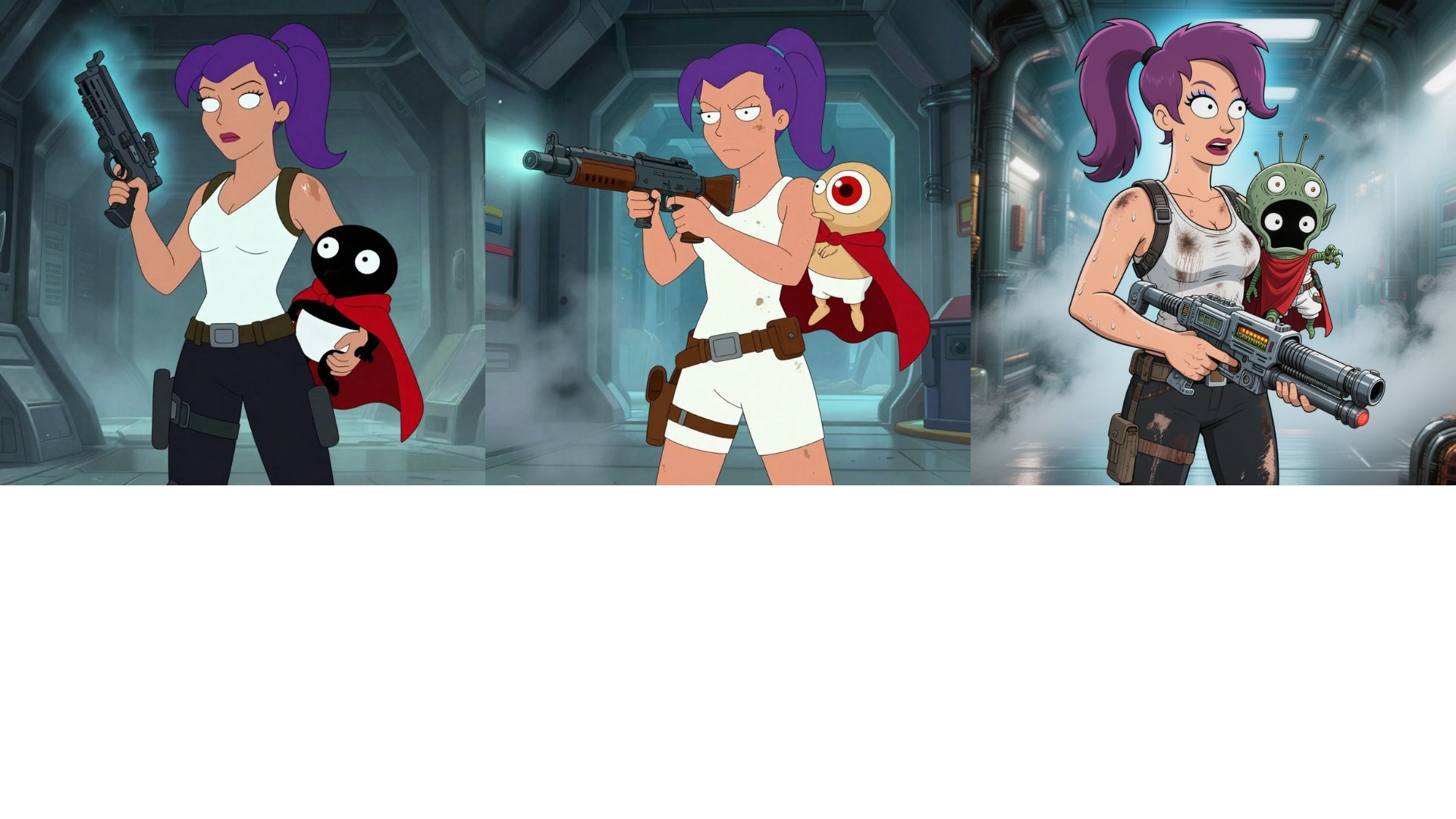}
    \vspace{-0.3em}
    \caption{Baseline vs CRAFT vs Hunyuan Image 3 (see
        Appendix~\ref{app:cost_prompts}, cost\_prompt3).}
    \label{fig:cost3_app}
\end{figure}

\begin{figure}[!h]
    \centering
    \includegraphics[width=0.65\textwidth]{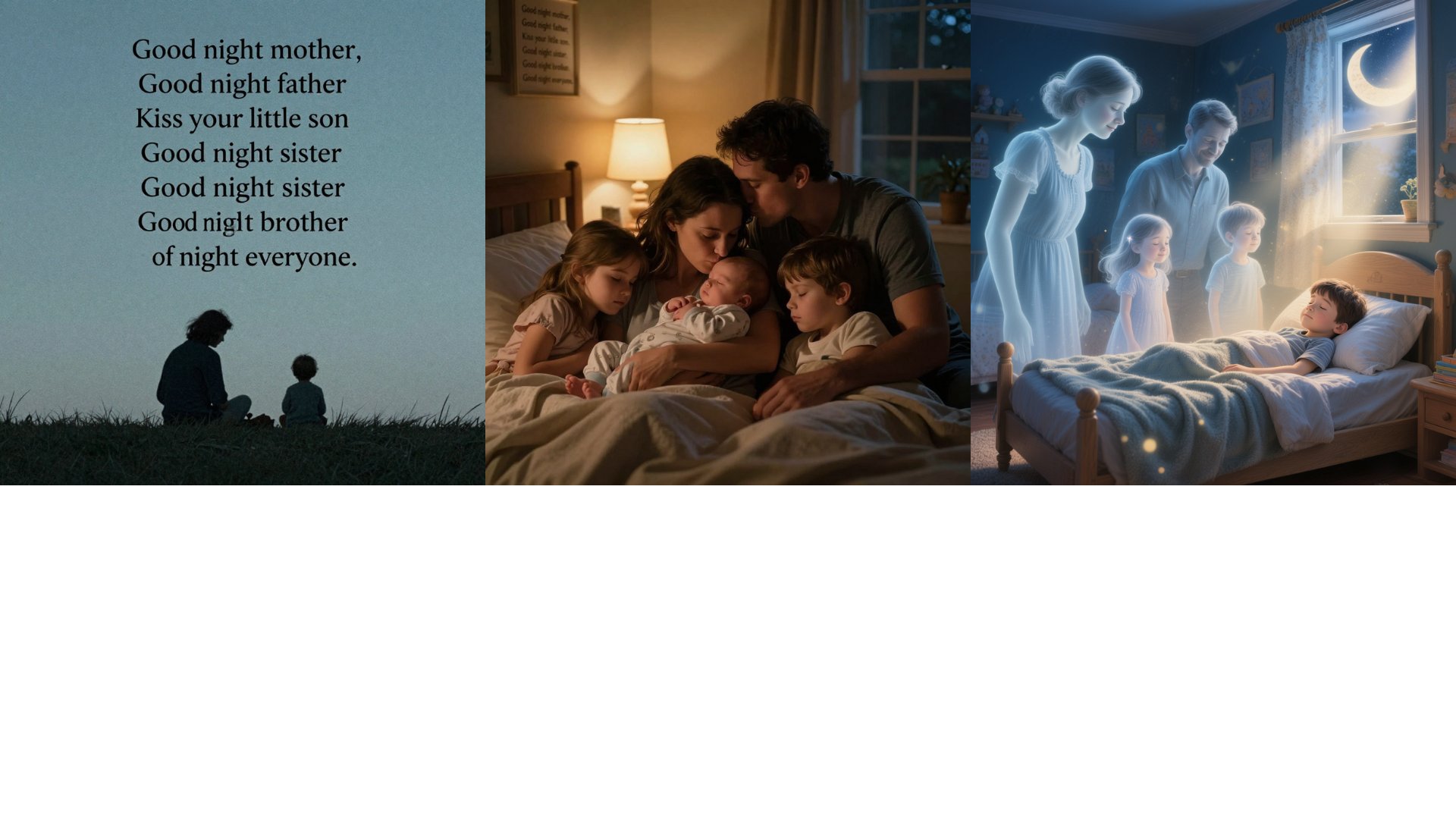}
    \vspace{-0.3em}
    \caption{Baseline vs CRAFT vs Hunyuan Image 3 (see
        Appendix~\ref{app:cost_prompts}, cost\_prompt4).}
    \label{fig:cost4_app}
\end{figure}

\begin{figure}[!h]
    \centering
    \includegraphics[width=0.65\textwidth]{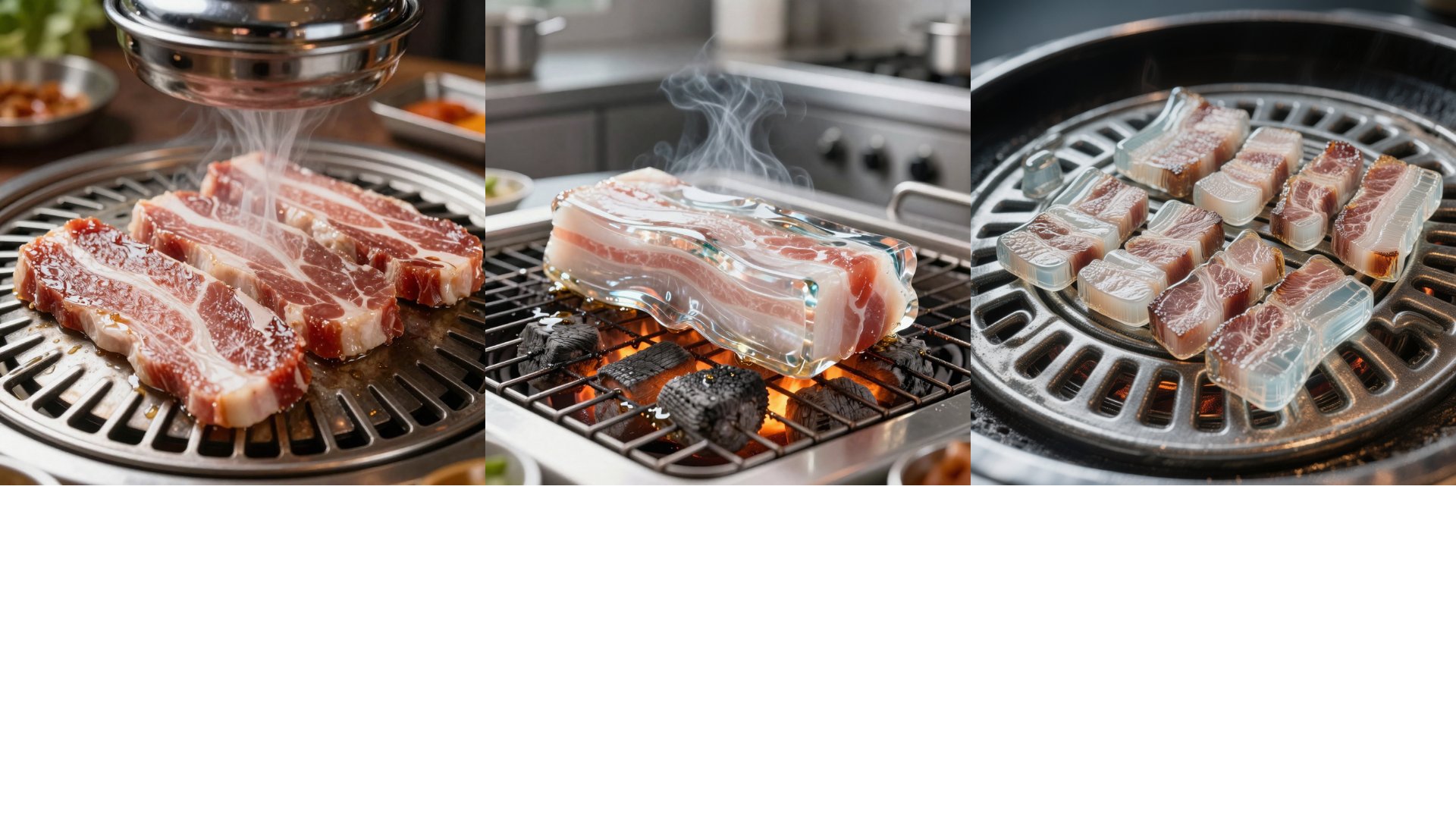}
    \vspace{-0.3em}
    \caption{Baseline vs CRAFT vs Hunyuan Image 3 (see
        Appendix~\ref{app:cost_prompts}, cost\_prompt5).}
    \label{fig:cost5_app}
\end{figure}

\begin{figure}[!h]
    \centering
    \includegraphics[width=0.65\textwidth]{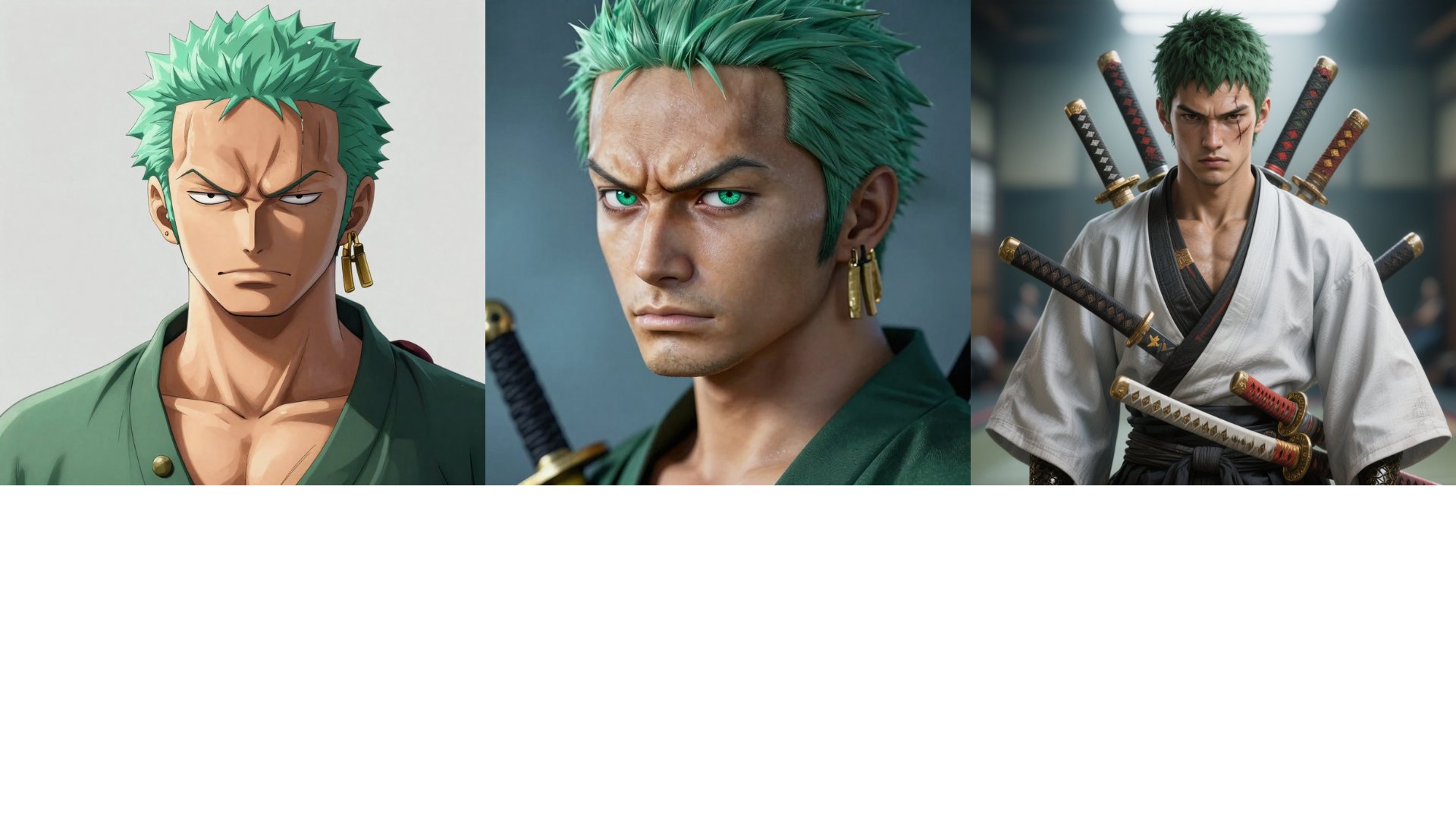}
    \vspace{-0.3em}
    \caption{Baseline vs CRAFT vs Hunyuan Image 3 (see
        Appendix~\ref{app:cost_prompts}, cost\_prompt6).}
    \label{fig:cost6_app}
\end{figure}

\end{document}